\documentclass{article}

\newcommand{\setformat}[1]{\def\conferenceformat{#1}}
\newcommand{\isformat}[1]{\ifnum\pdfstrcmp{\conferenceformat}{#1}=0}
\setformat{arxiv}

\usepackage{styles/iml_arxiv/arxiv_ml_institute}
\usepackage{algorithm}
\usepackage{algorithmic}
\usepackage{amsmath}
\usepackage{amssymb}
\usepackage{wrapfig}
\usepackage{float}
\usepackage{placeins} %

\usepackage[utf8]{inputenc} %
\usepackage[T1]{fontenc}    %
\usepackage{hyperref}       %
\usepackage{url}            %
\usepackage{booktabs}       %
\usepackage{amsfonts}       %
\usepackage{nicefrac}       %
\usepackage{microtype}      %
\usepackage{xcolor}         %
\usepackage{graphicx}       %
\usepackage{subcaption}
\usepackage{stmaryrd}

\usepackage{styles/iml_arxiv/ml_defs} %

\definecolor{mydarkblue}{rgb}{0,0.08,0.45}
\hypersetup{
    colorlinks=true,
    linkcolor=mydarkblue,
    citecolor=mydarkblue,
    filecolor=mydarkblue,
    urlcolor=mydarkblue,
}

\title{Diverse Topology Optimization \\ using Modulated Neural Fields}

\author{
  Andreas Radler$^{* \ 1}$ \qquad
  Eric Volkmann$^{* \ 1}$ \qquad
  \AND
  Johannes Brandstetter$^{\ 1,2}$ \qquad 
  Arturs Berzins$^{\ 1}$ \qquad \\ \\
{$^*$}{Equal contribution}\\ %
{$^1$}{LIT AI Lab, Institute for Machine Learning, JKU Linz, Austria}\\
{$^2$}{Emmi AI GmbH, Linz, Austria}\\
\texttt{\{radler, volkmann, brandstetter, berzins\}@ml.jku.at}
}

\begin{document}

\maketitle

\begin{abstract}

Topology optimization (TO) is a family of computational methods that derive near-optimal geometries from formal problem descriptions. 
Despite their success, established TO methods are limited to generating single solutions, restricting the exploration of alternative designs. 
To address this limitation, we introduce \emph{Topology Optimization using Modulated Neural Fields} (TOM) -- a data-free method that trains a neural network to generate structurally compliant shapes and explores diverse solutions through an explicit diversity constraint.
The network is trained with a solver-in-the-loop, optimizing the material distribution in each iteration. 
The trained model produces diverse shapes that closely adhere to the design requirements.
We validate TOM on 2D and 3D TO problems.
Our results show that TOM generates more diverse solutions than any previous method, all while maintaining near-optimality and without relying on a dataset.
These findings open new avenues for engineering and design, offering enhanced flexibility and innovation in structural optimization.
\footnote{The code is available at https://github.com/ml-jku/Topology-Optimization-Modulated-Neural-Fields}

\end{abstract}

\begin{figure}[ht!]
    \centering
    \begin{subfigure}[t]{0.19\linewidth}
        \centering
        \raisebox{0.45\height}{\includegraphics[height=0.1\textheight, keepaspectratio]{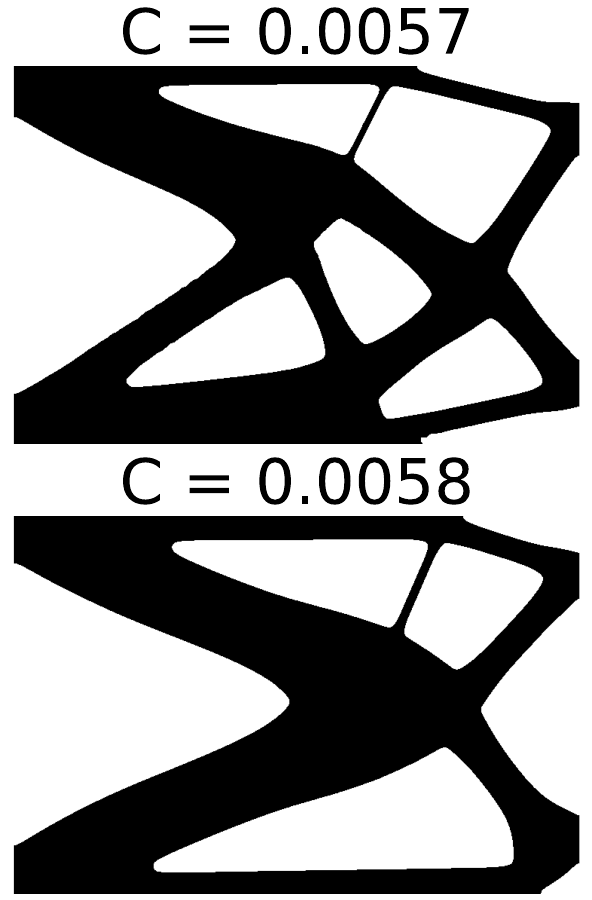}}
        \caption{Deflated barrier}
        \label{fig:DB_cantilever}
    \end{subfigure}
    \hfill
    \begin{subfigure}[t]{0.39\linewidth}
        \centering
        \includegraphics[height=0.2\textheight, keepaspectratio]{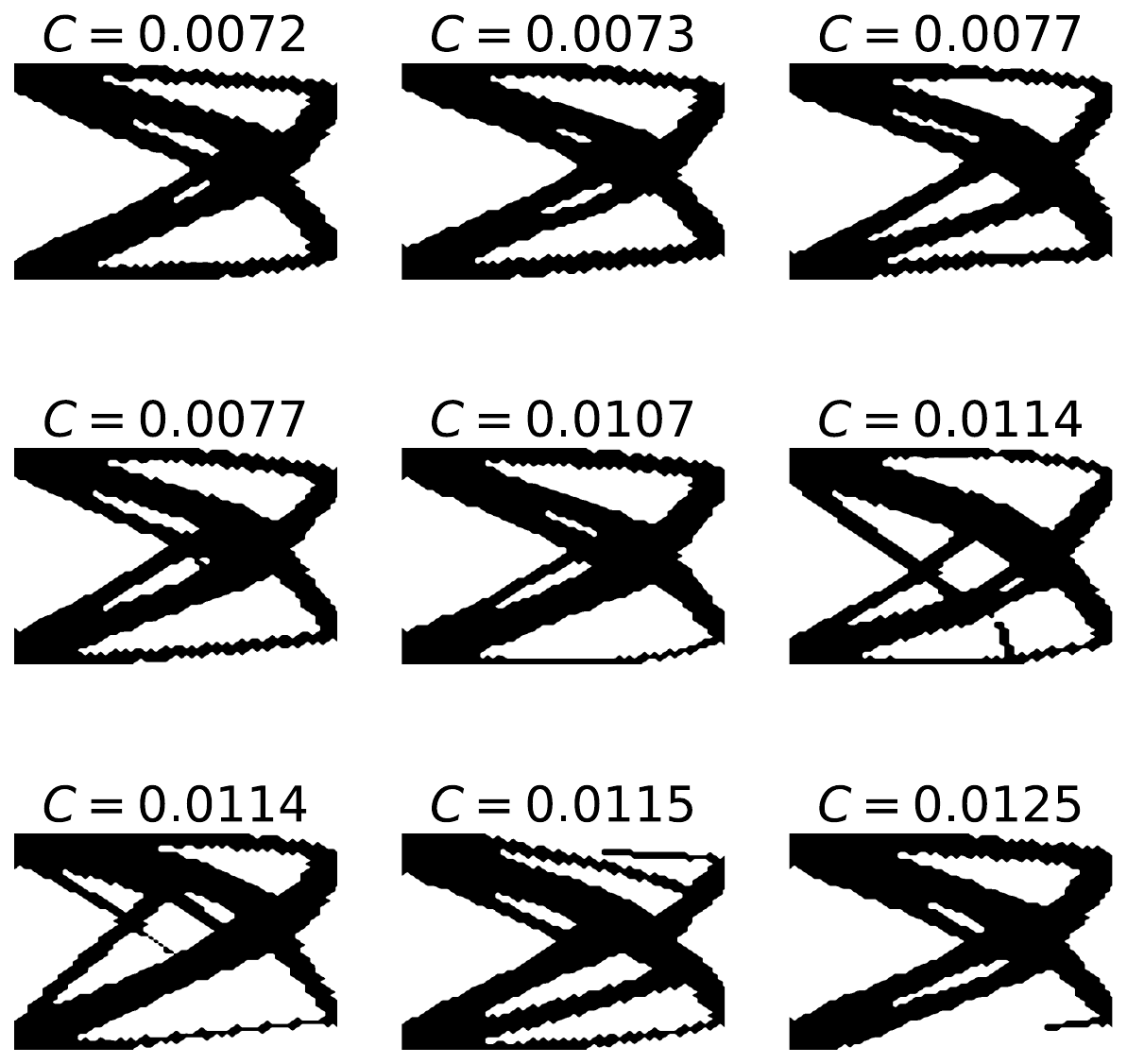}
        \caption{TopoDiff}
        \label{fig:difftopo_cantilever}
    \end{subfigure}
    \hfill
    \begin{subfigure}[t]{0.4\linewidth}
        \centering
        \includegraphics[height=0.2\textheight, keepaspectratio]{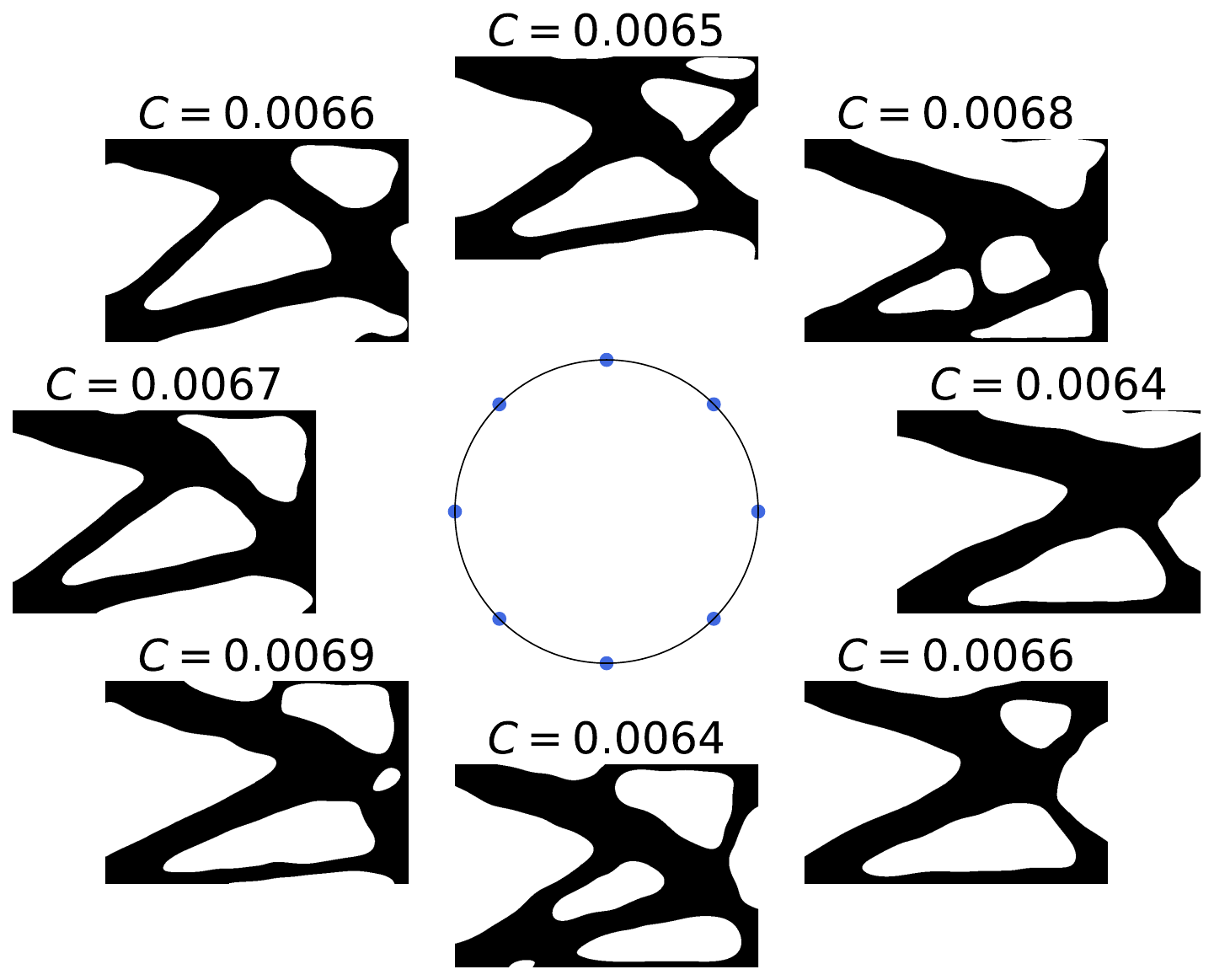}
        \caption{TOM (ours)}
        \label{fig:tom_cantilever}
    \end{subfigure}
    \caption{Solutions to the cantilever problem generated by three TO methods. (a) \emph{Deflated barrier} -- a classical TO method generating high-quality but few solutions. (b) \emph{TopoDiff} -- a data-driven diffusion model generating many, but low-quality solutions. (c) \emph{TOM} -- our modulated neural field trained with a solver-in-the-loop and a diversity constraint generating diverse near-optimal structures. Here, we use a circular modulation space to capture a smooth manifold of solutions.
    All methods minimize the structural compliance $C$, which measures the total displacement of the loaded shape.
    }
    \label{fig:results_cantilever}
\end{figure}

\section{Introduction}

Topology optimization (TO) is a computational design technique to determine the optimal material distribution within a given design space under prescribed boundary conditions. 
A common objective in TO is the minimization of structural compliance, also called strain energy, which measures the displacement under load and is inverse to the stiffness of the generated design.

Due to the non-convex nature of TO problems, these methods generally provide near-optimal solutions, with no guarantees of converging to global optima \citep{allaire2021shape}.
Despite this, TO has been a critical tool in engineering design since the late 1980s and continues to evolve with advances in computational techniques, including the application of machine learning.

Traditional TO methods %
produce a single design, which limits its utility. 
Engineering problems often require designs that balance performance with other considerations, such as manufacturability, cost, and aesthetics. These are associated with uncertainty, implicit knowledge, and subjectivity and are hard to optimize in a formal framework.
Generating multiple diverse solutions allows for exploring these trade-offs and provides more flexibility in choosing designs that meet both technical and real-world constraints. 
However, \emph{conventional TO methods} cannot efficiently explore and generate multiple high-quality designs, %
making it difficult to address this need. 
While \emph{data-driven methods} have shown potential in addressing this challenge, they rely on large training datasets \citep{topodiff_maze22, nie2021topologygan}, which are expensive and often infeasible to obtain in practice. 
Additionally, data-driven methods often fail to generalize beyond the training data distribution \citep{woldseth_use_2022}.

Therefore, we propose 
\textbf{T}opology \textbf{O}ptimization using \textbf{M}odulated
Neural Fields (TOM), an approach to 
generate diverse, near-optimal designs: A neural network parametrizes the shape representations and is trained to generate near-optimal solutions. The model is optimized using a solver-in-the-loop approach \citep{2020solver_in_the_loop}, where the neural network iteratively adjusts the design based on feedback from a physics-based solver. To enhance the diversity of solutions, we introduce an explicit diversity constraint during training, ensuring the network produces a range of solutions that still adhere to mechanical compliance objectives. TOM not only enables the generation of multiple, diverse designs but also leverages machine learning to explore the design space more efficiently than traditional TO methods \citep{sanu2024neural_good_bad_ugly}.

Empirically, we validate our method on TO for linear elasticity problems in 2D and 3D. 
Our results demonstrate that TOM obtains more diverse solutions than prior work while being substantially faster, remaining near-optimal, and circumventing adaptive mesh refinement by using continuous neural fields. 
This addresses a current limitation in TO and opens new avenues for automated engineering design.

Our main contributions are summarized as follows:
    
\begin{enumerate}
    \item We introduce TOM, the first method for data-free, solver-in-the-loop neural network training, which generates diverse solutions adhering to structural requirements.
    \item We introduce a novel diversity constraint variant for neural density fields based on the chamfer discrepancy that ensures the generation of distinct and meaningful shapes and enhances the exploration of the designs.
    \item Empirically, we demonstrate the efficacy and scalability of TOM on 2D and 3D problems, showcasing its ability to generate a variety of near-optimal designs.
    Our approach significantly outperforms existing methods in terms of solution diversity.

\end{enumerate}

\section{Background} 
\label{sec:preliminaries}

\subsection{Topology optimization}
\label{subsec:TO}
TO is a computational method developed in the late 1980s to determine optimal structural geometries from mathematical formulations \citep{BENDSOE1988197}. It is widely used in engineering to design efficient structures while minimizing material usage.
TO iteratively updates a material distribution within a design domain under specified loading and boundary conditions to enhance properties like stiffness and strength. 
Due to the non-convex nature of most TO problems, convergence to a global minimum is not guaranteed. Instead, the goal is to achieve a near-optimal solution, where the objective closely approximates the global optimum.
There are four prominent method families widely recognized in TO.
In this work, we focus on \emph{SIMP} and refer the reader to \citet{yago_topology_2022} for a more detailed introduction to TO.

\paragraph{Solid isotropic material with penalization (SIMP)} is a prominent TO method we adopt for TOM.
SIMP starts by defining a mesh with mesh points $\mathbf{x}_i \in \mathcal{X}, i \in \{1, ..., N \}$ in the design region.
The aim is to find a binary density function at each mesh point $\rho(\mathbf{x}_i) \in \{0, 1\}$, where $\rho(\mathbf{x}_i) = 0$ represents void and $\rho(\mathbf{x}_i) = 1$ represents solid material. 
To make this formulation differentiable, the material density $\rho$ is relaxed to continuous values in $[0, 1]$. 
\
A common objective of SIMP is to minimize the compliance $C$, which is a measure of deformation under load.
The SIMP objective is then formulated as a constrained optimization problem:
\begin{equation}
\begin{aligned}
\label{eq:simp}
\min : & \quad C(\rho) = \mathbf{u}^T \mathbf{K_\rho} \mathbf{u} \\
\text{s.t.} : & \quad V = \sum_{i=1}^N \mathbf{\rho}_i v_i \leq V^* \\
& \quad 0 \leq \mathbf{\rho}_i \leq 1 \quad \forall i \in N 
\end{aligned}
\end{equation}
where $\mathbf{u}$ is the displacement vector, $\mathbf{K}_\rho$ is the global stiffness matrix, $V$ is the shape volume, and $V^*$ is the target volume.
\
The density field is optimized iteratively.
In each iteration, a finite element (FEM) solver computes the compliance and provides gradients to update the density field $\rho$.
To encourage binary densities, intermediate values are penalized by raising $\rho$ to the power $p > 1$. 
Hence, the stiffness matrix is defined as $\mathbf{K}_\rho = \sum_{i=1}^{N} \rho_i^p \mathbf{K}_i $
, where $\mathbf{K}_i$ describes the stiffness of solid cells and depends on material properties.

\paragraph{Annealing} is employed to make the continuous relaxation closer to the underlying discrete problem \citep{Kirkpatrick1983Annealing}, enhancing the effectiveness of gradient-based optimization methods.
Annealing gradually adjusts the sharpness of a function during the optimization. 
For TO, this is often done by gradually increasing the penalty $p$ or by scheduling a sharpness filter.
A common choice is the Heaviside filter as defined in Equation~\ref{eq:heaviside} and further described in Appendix~\ref{subsec:density_filters}.

\paragraph{TO for multiple solutions.} 
Generating multiple solutions to a given TO problem 
is important to address real-world engineering challenges, since often considerations such as manufacturability and aesthetics influence the selection of the design. 
By generating diverse solutions, 
engineers can evaluate and compare alternative designs, ultimately selecting the most suitable option in a post-processing stage.
However, classical TO algorithms typically yield a single solution and do not ensure convergence to a global minimum \citep{multi_TO_Papadopoulos_2021}. 
\\
\citet{multi_TO_Papadopoulos_2021} introduce the deflated barrier (DB) method, an extension of the classical SIMP approach that can find multiple solutions. 
By employing a search strategy akin to depth-first search, DB identifies multiple solutions without relying on initial guess variations, thereby enhancing design diversity.

\subsection{Shape generation with neural networks}

\paragraph{Neural fields} offer a powerful framework for geometry processing, utilizing neural networks to model shapes implicitly.
Unlike conventional explicit representations like meshes or point clouds, implicit shapes are defined as level sets of continuous functions.
This approach enables high-quality and topologically flexible shape parameterizations \citep{chen2019implicit}. The two prevalent methods for representing implicit shapes are signed distance functions (SDF) \citep{park2019deepsdf, Atzmon2020sal} and density (or occupancy) \citep{mescheder2019occupancy} fields. We opt for the density representation due to its compatibility with SIMP optimization.
\\
Given a $d_x \in \{2, 3\}$ dimensional domain $\mathcal{X} \subset \mathbb{R}^{d_x}$, a neural density field employs a neural network $f_\theta: \mathcal{X} \to (0, 1)$ with parameters $\theta$ to define the shape $\Omega := \{x \in \mathcal{X} | f_\theta(x) \geq \tau\}$ as the $\tau \in [0,1]$ super-level-set of $f_{\theta}$.

\paragraph{Conditional neural fields.}
While a neural density field represents a single shape, a \emph{conditional} neural field represents a set of shapes with a single neural network \citep{Mehta21modulation}.
This can be regarded as a lossy compression of multiple shapes into a single neural network \cite{park2019deepsdf}. 
\\
Generally, one can condition on text, point clouds, or other modalities of interest \cite{zhang_3dshape2vecset_2023}.
In this work, we use a modulation code $\mathbf{z} \in \mathbb{R}^{d_z}$ as an additional input to the network.
The resulting network $f_\theta(\mathbf{x}, \mathbf{z})$ parameterizes a set of shapes.
There are different ways to incorporate the modulation vector into the network, such as input concatenation \citep{park2019deepsdf}, hypernetworks \citep{Ha17hypernetworks}, or attention \citep{Rebain2022Attention}.
In this work, we use input concatenation, as it is simple and fast to train.

\paragraph{Diversity constraint} can be used to modify the TO problem formulation \eqref{eq:simp} to facilitate the discovery of multiple solutions.
The diversity constraint introduced in geometry-informed neural networks (GINNs) \citep{GINNs} defines a diversity measure $\delta$ on the set of shapes $\{ \Omega_i \}$ as
\begin{align}
    \label{eq:div}
    \delta (\{ \Omega_i \}) = \Bigl( \sum_j \bigl[ \min_{k\neq j} d(\Omega_j, \Omega_k) \bigr]^{1/2} \Bigr)^2 \ .
\end{align}
This measure builds upon a chosen dissimilarity function $d(\Omega_i, \Omega_j)$.
Essentially, $\delta$ encourages diversity by maximizing the distance between each shape and its nearest neighbor.
GINNs utilize a dissimilarity function defined on the boundary, however, their approach only works for SDFs.
Therefore, we adjust the chamfer discrepancy (Equation \ref{eq:1_sided_chamfer}) as an alternative dissimilarity function tailored for diversity on density fields in Section \ref{sec:diversity}.

\subsection{Topology optimization with neural networks}

\begin{figure*}[t!]
    \centering
    \isformat{icml}
        \includegraphics[height=0.21\textwidth]{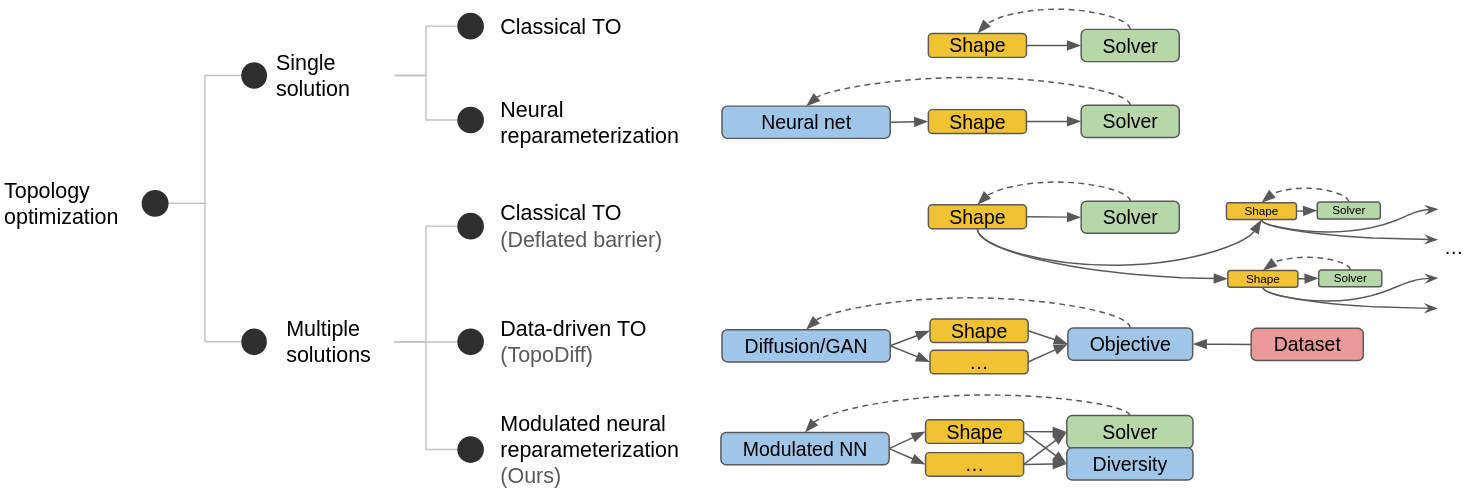}
    \else
        \includegraphics[width=\textwidth]{figures/taxonomy/TO_classification_dataset_right.png}
    \fi
    \caption{Taxonomy of classical and neural topology optimization methods. The dashed lines indicate iterative updates, such as gradient descent. TOM is the first data-free method to produce multiple shapes with a neural network. 
    }
    \label{fig:TO_classification}
\end{figure*}

Figure \ref{fig:TO_classification} presents an overview of various TO methods that search for either single or multiple solutions, further categorized by their use of neural networks.

\paragraph{Neural reparameterization}
uses a neural network to represent the material distribution in a discretization-free manner. 
Existing work explores training a single shape by parametrizing the material \emph{density} \citep{sosnovik_neural_2017, Bujny_TO_reparametrization, Chandrasekhar21tounn, Hoyer2019NeuralRI} or \emph{boundary} \citep{deng2021_TO_levelset}, which is optimized with a solver-in-the-loop.
\\
NTopo \citep{Zehnder21ntopo} is conceptually related to TOM, but employs a conditional neural field to parameterize individual solutions across multiple TO problems. 
Specifically, NTopo conditions the neural field on factors such as target volume or force vector position, producing a unique solution for each configuration.
In contrast, TOM extends this idea by enabling the discovery of multiple solutions for a single configuration, rather than restricting each configuration to a single solution.

\paragraph{Data-driven neural TO}
uses a dataset of generated solutions to train neural networks that can then generate diverse solutions.
Most prior works use generative adversarial networks (GANs) \citep{goodfellow_GAN, gillhofer2019gan, wang2020topogan, nie2021topologygan}.
As GANs are often exhibit mode collapse \citep{Che17mode, wang2020topogan}, more recent work showed the superiority of diffusion models \cite{sohl2015deep, ho2020denoising} for data-driven TO \citep{topodiff_maze22}.
An important weakness of data-driven methods is the large amount of training data, often in the tens of thousands for simple 2D problems \citep{topodiff_maze22, nie2021topologygan}.
\citet{woldseth_use_2022} explicitly criticize data-driven methods as these ``produce poor designs, are expensive to obtain, and are very restricted in terms of the variety of problems and mesh resolutions".

\section{Method}
\label{sec:method}

\subsection{Topology optimization using modulated neural fields}

\begin{figure*}[t!]
    \centering
    \isformat{icml}
        \includegraphics[height=0.21\textwidth]{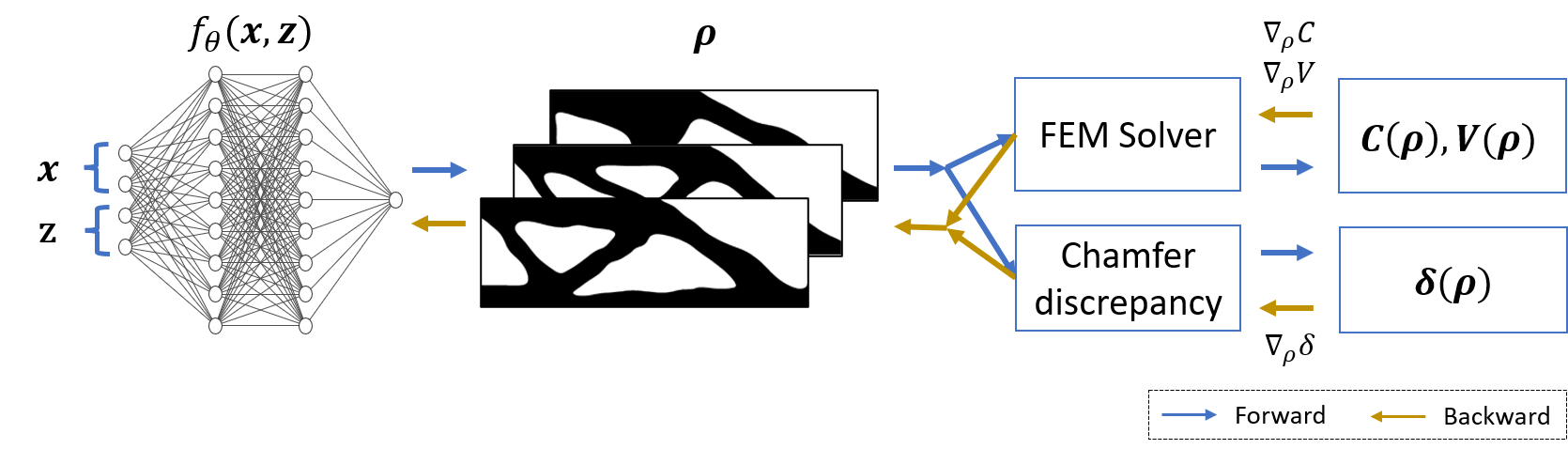}
    \else
        \includegraphics[width=\textwidth]{figures/method/Method-wider_bottom2.png}
    \fi
    \caption{A single iteration of TOM trained with $M=3$ shapes in parallel.
    In each iteration, the input to the network consists of the mesh vertices $\{ \mathbf{x}_i \}_{i = 1, ...,  N}$ and modulation vectors $\{ \mathbf{z} \}_{j = 1, ..., M}$.
    The network outputs densities $\rho_j$ at the vertices $\mathbf{x}_i $ for each shape.
    The densities are passed to the FEM solver which computes the compliances $C_j$ and volumes $V_j$, as well as their gradients $\nabla_\rho C_j$ and $\nabla_\rho V_j$.
    The diversity loss is based on the chamfer discrepancy between the surface points of the shapes.
    These pairwise discrepancies are used to compute the diversity loss $\delta(\rho)$ and its gradient $\nabla_\rho \delta$.
    }
    \label{fig:placeholder}
\end{figure*}

\paragraph{Definitions.}
Let $\mathcal{X} \subset \mathbb{R}^{d_x}$
be the domain of interest in which there is a shape $\Omega \subset \mathcal{X}$. 
Let $\mathcal{Z} \subset \mathbb{R}^{d_z}$ be a discrete or continuous modulation space, where each $\mathbf{z} \in \mathcal{Z}$ parametrizes a shape $\Omega_\mathbf{z}$. The possibly infinite set of all shapes is denoted by $\Omega_\mathcal{Z} = \{ \Omega_\mathbf{z} | \mathbf{z} \in \mathcal{Z} \}$.
The modulation vectors $\{\mathbf{z}_{i} \}$ are sampled according to a probability distribution $p(\mathcal{Z})$.
\newline
For density representations a shape $\Omega_\mathbf{z}$ is defined as the set of points with a density greater than the level $\tau \in [0,1]$, formally $\Omega_\mathbf{z} = \{\mathbf{x} \in \mathcal{X} \mid \rho_\mathbf{z}(\mathbf{x}) > \tau \}$.
While some prior work on occupancy networks \citep{Occupancy_Networks} treats $\tau$ as a tunable hyperparameter, we follow \citet{fenitop} and fix the level at $\tau = 0.5$. 
We model the density $\rho_\mathbf{z}(\mathbf{x}) = f_\theta(\mathbf{x}, \mathbf{z})$, corresponding to the modulation vector $\mathbf{z}$ at a point $\mathbf{x}$ using a neural network $f_\theta$, with $\theta$ being the learnable parameters.

\paragraph{TO using modulated neural fields.}
Our approach, akin to standard SIMP, formulates a constrained optimization problem aiming to minimize an objective function for multiple shapes simultaneously, while adhering to a volume constraint for each shape. 
In the context of linear elasticity, the objective is to minimize compliance, which measures deformation under load.
Central to our method is the introduction of a diversity constraint, denoted as $\delta(\Omega_\mathcal{Z})$, which is defined over multiple shapes to reduce their similarity.
This leads to the following constrained optimization problem:

\begin{equation}
\begin{aligned}
\label{eq:TOM}
\min : & \quad \mathbb{E}_{\mathbf{z} \sim p(\mathcal{Z})} \left[ C(\rho_{\mathbf{z}}) \right]
\\
\text{s.t.} : & \quad V_{\mathbf{z}} = \int_\mathcal{X} \mathbf{\rho}_{\mathbf{z}}(\mathbf{x}) d\mathbf{x} \leq V^*\\
& \quad 0 \leq \mathbf{\rho}_{\mathbf{z}}(\mathbf{x}) \leq 1 \\
& \quad \delta^* \leq \delta(\Omega_\mathcal{Z})
\end{aligned}
\end{equation}

where $V^*$ and $\delta^*$ are target volumes and diversities, respectively.
TOM updates the density fields of multiple shapes iteratively. In each iteration, the density distribution of each shape is computed and the resulting densities are passed to the FEM solver.
The FEM solver calculates the compliance loss $C$ and gradients $\nabla C$.
To accelerate the computation, parallelization of the FEM solver is employed across multiple CPU cores, with a separate process dedicated to each shape.
\\ 
The diversity loss and its gradient (see Section \ref{sec:diversity}) is computed on the GPU using PyTorch \citep{paszke2017pytorch}, along with any optional geometric losses similar to GINNs (see Section \ref{sec:geom_constraints}).
\\
We use the adaptive augmented Lagrangian method (ALM) \citep{basir2023adaptive} to automatically balance multiple loss terms. 
Additionally, we gradually increase the sharpness parameter $\beta$ of the Heaviside filter (Equation \ref{eq:heaviside}) which acts as annealing. 
The complete TOM method is concisely presented in Algorithm \ref{alg:TOM}.

\subsection{Diversity}
\label{sec:diversity}

The diversity loss, defined in Equation \ref{eq:div}, requires the definition of a dissimilarity measure between a pair of shapes. 
Generally, the dissimilarity between two shapes can be based on either the volume of a shape $\Omega$ or its boundary $\partial \Omega$.
GINNs \citep{GINNs} use boundary dissimilarity, which is easily optimized for SDFs since the value at a point is the distance to the zero level set. However, this dissimilarity measure is not applicable to our density field shape representation.
Hence, we propose a boundary dissimilarity based on the chamfer discrepancy.
We also develope a volume-based dissimilarity (detailed in Appendix \ref{subsec:volume_dissimilarity}), however, we focus on boundary diversity due to its promising results in early experiments.

\paragraph{Diversity on the boundary via differentiable chamfer discrepancy.}
To define the dissimilarity on the boundaries of a pair of shapes $(\partial \Omega_1, \partial \Omega_2)$, we use the one-sided chamfer discrepancy (CD):
\begin{align}
\label{eq:1_sided_chamfer}
    \text{CD}(\partial \Omega_1, \partial \Omega_2) = \frac{1}{\left| \partial \Omega_1 \right|} \sum_{x \in \partial \Omega_1} \min_{\Tilde{x} \in \partial \Omega_2} ||x-\Tilde{x}||_2
\end{align}
where $x$ and $\Tilde{x}$ are sampled points on the boundaries.
To use the CD as a loss, it must be differentiable w.r.t. the network parameters $\theta$.
However, the chamfer discrepancy $\text{CD}(\partial\Omega_1, \partial\Omega_2)$ depends only on the boundary points $x_i \in \partial \Omega$ which only depend on $f_\theta (x_i)$ implicitly.
Akin to prior work \citep{chen_converting_2010, Berzins2023bs, mehta_level_2022}, we apply the chain rule and use the level-set equation to derive
\isformat{icml}

    \begin{align}
        \frac{\partial \text{CD}}{\partial \theta} &= \frac{\partial \text{CD}}{\partial x} \frac{\partial x}{\partial y} \frac{\partial y}{\partial \theta} \nonumber \\ 
        &= \frac{\partial \text{CD}}{\partial x} \frac{\nabla_x f_\theta}{|\nabla_x f_\theta|^2} \frac{\partial y}{\partial \theta} \label{eq:level_set}
    \end{align}
\else
    \begin{align}
    \label{eq:level_set}
        \frac{\partial \text{CD}}{\partial \theta} = \frac{\partial \text{CD}}{\partial x} \frac{\partial x}{\partial y} \frac{\partial y}{\partial \theta} = \frac{\partial \text{CD}}{\partial x} \frac{\nabla_x f_\theta}{|\nabla_x f_\theta|^2} \frac{\partial y}{\partial \theta} 
    \end{align}
\fi
where $y = \rho$ is the density field in our case.
We detail this derivation in Appendix \ref{subsec:chamfer_diversity}.

\paragraph{Finding surface points on density fields} is substantially harder than for SDFs.
For an SDF $f$, surface points can be obtained by flowing randomly initialized points along the gradient $\nabla f$ to the boundary $f=0$. 
However, density fields $g$ generally do not satisfy the eikonal equation as $|\nabla g| \neq 1$, causing gradient flows to get stuck in local minima. 
Instead, we employ a robust algorithm that relies on dense sampling and binary search, as detailed in Algorithm \ref{alg:boundary_points}.

\subsection{Formalizing geometric constraints}
\label{sec:geom_constraints}

The compliance and volume losses are computed by the FEM solver on a discrete grid.
In addition, we use geometric constraints similar to GINNs, leveraging the continuous field representation. 
This enables learning finer details, especially at the interfaces. 
While these constraints can also be enforced via the TO framework by assigning solid regions ($\rho = 1$ values) to specific mesh cells, this would require a higher mesh resolution, increasing the computational cost of the solver. 
Instead, we adapt the geometric constraints of GINNs to operate with a 
density representation.
The applied constraints and further details are in Table \ref{tab:constraints} and Appendix \ref{app:constraints}. 
Crucially, these are computed in parallel on the GPU, accelerating the optimization process.

\section{Experiments}

We solve three standard TO benchmark problems comparing TOM to established classical and data-driven methods (see Section \ref{sec:baselines}).
For all our TOM experiments, we employ the WIRE architecture \citep{saragadam2023wire}, which uses wavelets as activation functions.
This imposes an inductive bias towards high-frequencies, while being more localized than, e.g., a sine activation \citep{sitzmann2019siren}.
We use a 2-dimensional modulation space and sample vectors uniformly on a circle with radius $r$. 
We denote a circle centered at the origin with radius r as the 1-sphere $\mathcal{S}^1(r) = \{\mathbf{x} \in \mathbb{R}^2 : \|\mathbf{x}\| = r\}$.
This 1D manifold embedded in the 2D modulation space assures that modulation vectors are sufficiently far apart from each other to avoid mode collapse.
The radius controls the initial diversity of the shapes and is an important hyperparameter to tune (see also Appendix \ref{app:ablations}).
At each iteration, the number of shapes $M$ is either 9 or 25, depending on the problem. 
Further experimental details and hyperparameters can be found in Appendix \ref{app:exp_details}.

\subsection{Problem definitions}

We apply our method to common linear elasticity problems \citep{sigmund_morphology-based_2007,fenitop} in two and three dimensions, namely the Messerschmitt-Bölkow-Blohm (MBB) beam, the cantilever beam, and the jet engine bracket.
Due to the symmetry of the MBB beam, we follow \citet{multi_TO_Papadopoulos_2021} and optimize only the right half.
Detailed descriptions of these problems are provided in Appendix \ref{app:problem_defs} and Figure \ref{fig:problem_defs}.

\begin{figure*}[ht!]
\centering
\begin{subfigure}[t]{0.24\textwidth}
    \centering

    \hspace*{-0.5cm}
    \raisebox{0.45cm}{
    \includegraphics[height=1.5cm]{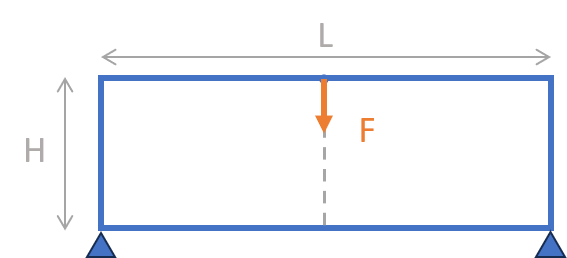}
    }
    \caption{MBB}
    \label{fig:beam_def}
\end{subfigure}
\hfill
\begin{subfigure}[t]{0.24\textwidth}
    \centering
    \hspace*{-0.3cm}
    \raisebox{0.2cm}{
    \includegraphics[height=2.2cm]{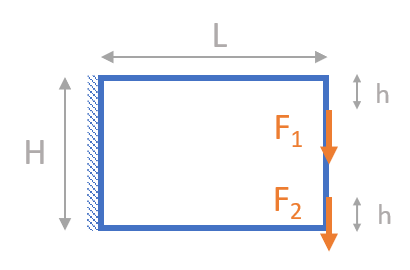}
    }
    \caption{Cantilever}
    \label{fig:cantilever_def}
\end{subfigure}
\hfill
\begin{subfigure}[t]{0.48\textwidth}
    \centering
    \raisebox{0.3cm}{
    \begin{subfigure}[t]{0.48\textwidth}
        \centering
        \includegraphics[height=2.2cm]{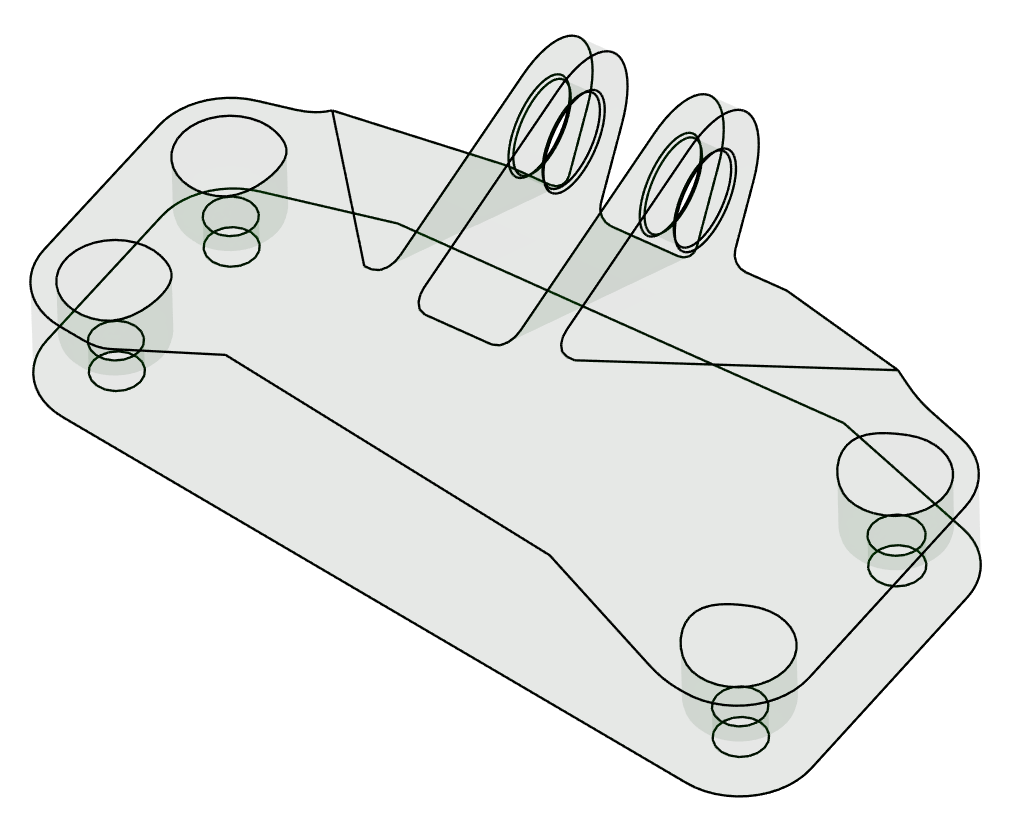}
        \label{fig:simjeb_env}
    \end{subfigure}
    \hfill
    \begin{subfigure}[t]{0.48\textwidth}
        \centering
        \raisebox{0.2cm}{
        \includegraphics[height=2.1cm]{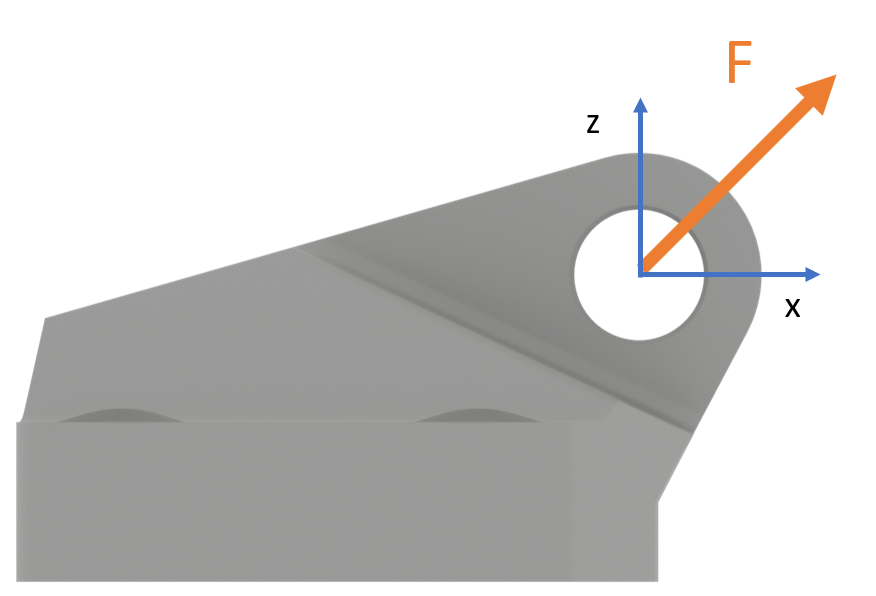}
        }
        \label{fig:simjeb_def}
    \end{subfigure}
    }
    \caption{Jet engine bracket}
    \label{fig:simjeb_combined}
\end{subfigure}

\caption{Three benchmark problem definitions.
(a) \emph{MBB}: The boundary is fixed at the attachment points at the bottom left and right corners. The dotted line indicates the symmetry axis at $\frac{L}{2}$ where the force $F$ is applied.
(b) \emph{Cantilever}: Forces $F_1$, $F_2$ are applied at distance $h$ from the upper and lower boundaries. The left boundary is fixed.
(c) \emph{Jet engine bracket}. Left: Available design region with six cylindrical interfaces where the shape must attach.
Right: A diagonal force is applied at the two central interfaces. The bracket is fixed at the four side interfaces.
}
\label{fig:problem_defs}
\end{figure*}

\subsection{Baselines}
\label{sec:baselines}

The experiments compare TOM with state-of-the-art classical and data-driven approaches (see Figure \ref{fig:TO_classification}).
As introduced in Section \ref{subsec:TO}, the standard method for single shape TO is SIMP \citep{BENDSOE1988197}, for which we use the FeniTop implementation \citep{fenitop} -- a well-documented Python wrapper of the standard FenicsX/DolfinX FEM solvers \citep{baratta2023dolfinx}.

\paragraph{Deflated barrier} (DB) method \citep{multi_TO_Papadopoulos_2021} is the state-of-the-art method for finding multiple solutions to TO problems.
DB is a sequential algorithm that cannot perform multiple solver steps in parallel, as detailed in Section \ref{sec:preliminaries}.
For the jet engine bracket, a comparison to DB is omitted, given DB's slow performance (see Section \ref{sec:results}), complexity, and hyperparameter sensitivity.

\paragraph{TopoDiff} \citep{topodiff_maze22} is a state-of-the-art data-driven TO solver. It is a diffusion model trained on 33,000 TO problem-solution pairs, discretized as $64\times64$ images.
Similar to DB, we could not apply TopoDiff to the 3D benchmark problem, since neither a pretrained model nor a large 3D TO dataset is publicly available, and generating one ourselves is infeasible.

\subsection{Metrics}

\paragraph{Quality.}
To evaluate structural characteristics, we report the mean and standard deviations for compliance $C$ and volume $V$. 
To further quantify the dispersion of compliance among the samples, we report the minimum and maximum compliances.
We use the FenicsX numerical solver \citep{baratta2023dolfinx} to compute the compliances and volumes.
For each problem, we use the same high-resolution mesh across all methods to compute the evaluation metrics.
This minimizes the discretization error and prevents numerical artifacts due to the mesh dependency of the solver.
\\
Following \citet{topodiff_maze22}, we also compute the \emph{load violation} $\text{LV} : \Omega \mapsto \{0,1\}$ for each solution $\Omega$. LV is 1 if the shape has no material anywhere the load applies, which necessarily leads to a high compliance, and 0 otherwise.
For each method, we report the LV \emph{ratio} of all produced solutions (LVR) and filter these to report the other metrics.

\paragraph{Diversity.}
As a diversity measure, we use the Hill numbers $D_q(S), q \in \mathbb{R}$ over a set $S$ as introduced by \citet{leinster2024entropy}.
For $q=2$, the Hill number corresponds to the expected dissimilarity if two elements of a set are sampled with replacement.
As a dissimilarity metric we choose the Wasserstein distance, as it is a mathematically well-defined distance between distributions.
Further, we use the sliced-1 Wasserstein distance approximation, a computationally cheap implementation \citep{flamary2021pot}.
Hence, computing the Hill number $D_2(\Omega_\mathcal{Z})$ corresponds to computing the expected Wasserstein distance as follows:
\begin{equation}
    \mathbb{E}[W_1] \coloneqq \mathbb{E}_{\mathbf{z}_i, \mathbf{z}_j \sim p(\mathcal{Z})} \bigl[ W_1(\Omega_{\mathbf{z}_i}, \Omega_{\mathbf{z}_j}) \bigr] .
\end{equation}

\paragraph{Computational cost.}
We report the wall clock time in Table \ref{tab:results}, to give a sense of the practicality of running each method. 
Additionally, we further characterize the computational cost of different methods in Table \ref{tab:additional_results} by reporting the number of solver steps, since for large meshes (e.g., more than 10K elements), this becomes the main computational expense. For this reason, we also report the mesh size. Lastly, we report the number of iterations of the network (i.e., optimizer steps), as these account for the parallelization of TOM.

\begin{figure}[t!]
    \centering
    \begin{subfigure}[t]{0.19\linewidth}
        \centering
        \raisebox{0.7\height}{\includegraphics[height=0.06\textheight, keepaspectratio]{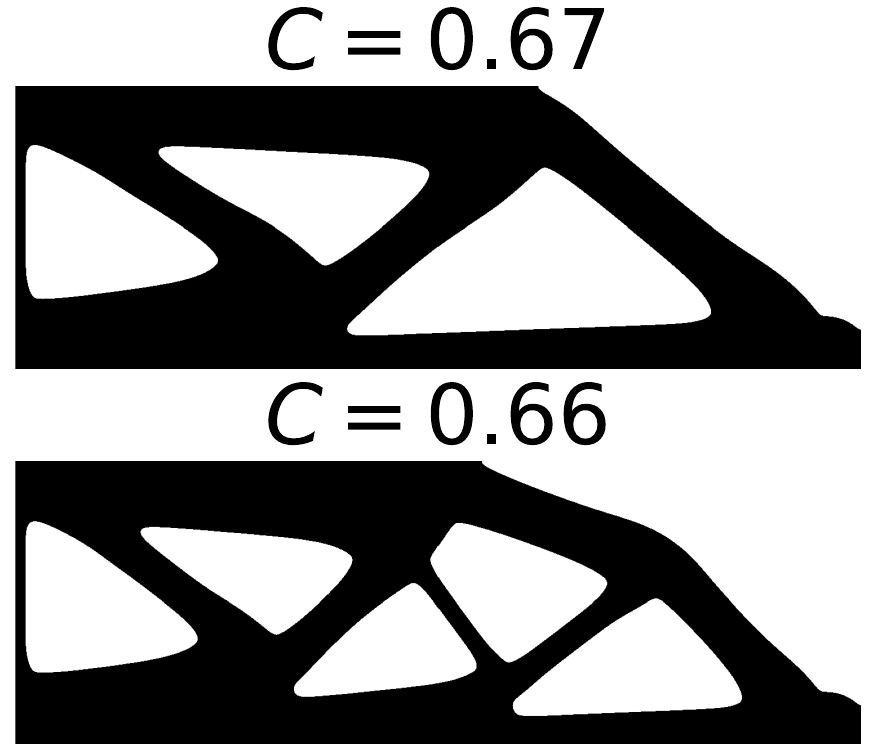}}
        \caption{Deflated barrier}
        \label{fig:DB_mbb}
    \end{subfigure}
    \hfill
    \begin{subfigure}[t]{0.39\linewidth}
        \centering
        \raisebox{0.17\height}{
        \includegraphics[height=0.12\textheight, keepaspectratio]{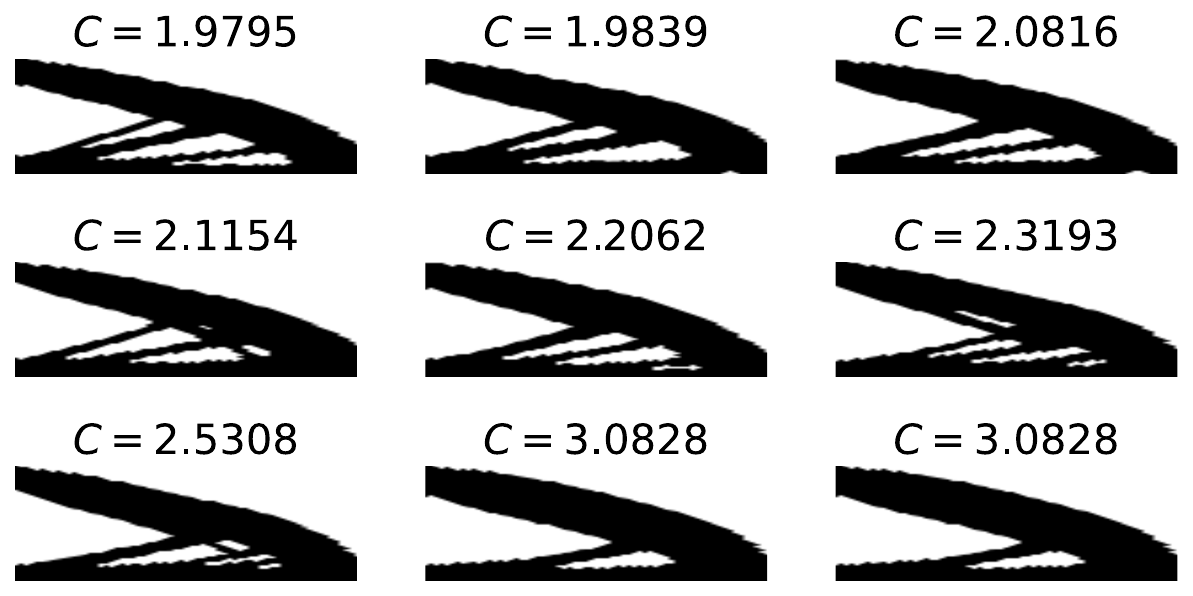}
        }
        \caption{TopoDiff}
        \label{fig:difftopo_mbb}
    \end{subfigure}
    \hfill
    \begin{subfigure}[t]{0.4\linewidth}
        \centering
        \includegraphics[height=0.16\textheight, keepaspectratio]{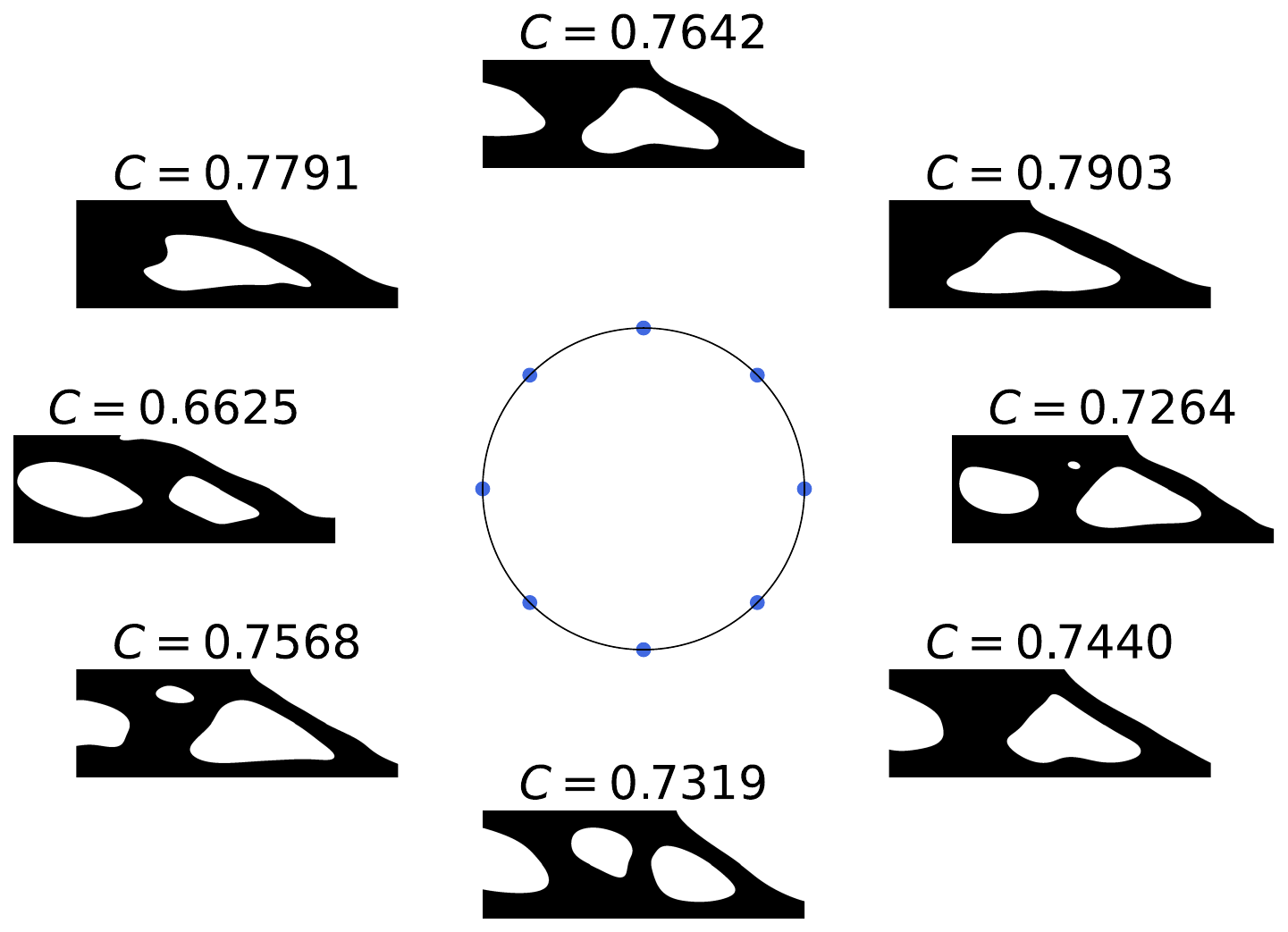}
        \caption{TOM (ours)}
        \label{fig:tom_mbb}
    \end{subfigure}
    \caption{
    Solutions to the \emph{MBB beam} problem generated by three TO methods.
    }
    \label{fig:results_beam}
\end{figure}

\subsection{Results}
\label{sec:results}

\begin{wrapfigure}[19]{r}{0.42\textwidth} %
\centering
\vspace{-34pt} %
\includegraphics[width=0.4\textwidth]{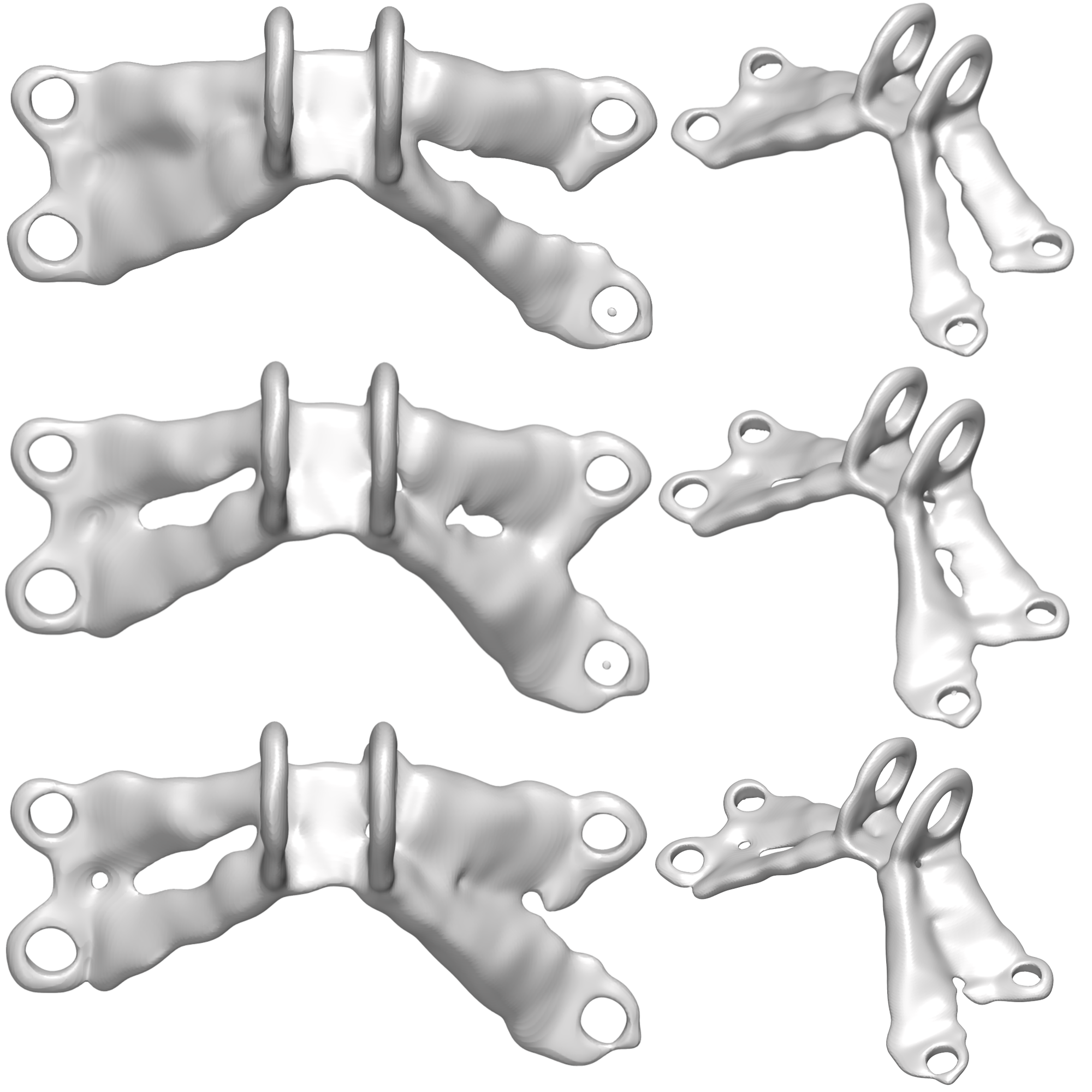}
\caption{
    Three different jet engine bracket designs generated by TOM. Left: top view. Right: isometric view. Notably, all three generated designs show a similar low compliance as the single solution by the FeniTop baseline. These explored solutions can be further refined and post-processed by a classical pipeline.
}
\label{fig:results_simjeb}
\end{wrapfigure}

The quantitative results of our experiments are summarized in Table \ref{tab:results}.
Qualitatively, we show different results in Figures \ref{fig:results_cantilever}, \ref{fig:results_beam},  and \ref{fig:results_simjeb}. 
Further quantitative and qualitative results are shown in Appendix \ref{app:ablations}.

\paragraph{TOM has high quality and diversity.}
Across all experiments, TOM finds diverse solutions with a near-optimal compliance.
DB has better compliance but a substantially lower diversity.
TopoDiff solutions are similarly diverse, but have poor compliance. For the cantilever, we filtered $\text{LVR}=87\%$ of TopoDiff outputs due to load violations, whereas for TOM $8\%$.
Surprisingly, the high LVR of TopoDiff contradicts the $\text{LVR}=0\%$ reported by \citet{topodiff_maze22}.
We attribute this to the cantilever being out of distribution from TopoDiff's training dataset.

\paragraph{TOM outperforms TopoDiff.}
At a first glance, the TopoDiff results in Figure \ref{fig:results_cantilever} (b) may appear visually appealing -- the shapes exhibit thin features and horizontal symmetry. However, the quantitative examination reveals that TopoDiff produces shapes with poor compliance and high LVR, discussed further in Appendix \ref{app:LV}. 
In contrast, TOM achieves lower compliance, lower LVR, and greater diversity. It is also data-free, eliminating the need for an expensive pretraining dataset.
Furthermore, TopoDiff requires around 9 seconds for inference. Although training time is not reported, it is expected to be significantly higher due to the complex architecture, conditioning and data, as well as training across problem instances, limiting the direct comparability of speed.

\begin{table*}[!t]
    \caption{Results on the three benchmark problems.
   $C$, $\min(C)$, and $\max(C)$
   are the mean ($\pm$ standard deviation), minimum, and maximum compliances.
   LVR is the ratio of load-violating solutions.
   As these always result in high compliance, results marked with a * report statistics after filtering such solutions.
   $V$ is the mean ($\pm$ standard deviation) of the volumes as a fraction of the available volume.
   $E(W_1)$ quantifies the diversity as the expected Wasserstein-1 distance between solutions.
   $N$ is the number of solutions for methods that produce a finite number of samples.
   TopoDiff and TOM produce a continuous distribution that can be sampled.
   Last is the wall-clock time to run the methods.
   Table \ref{tab:additional_results} contains further computational characterization. 
   }
    \centering
    \LARGE
    \resizebox{\textwidth}{!}{
\begin{tabular}{llcccccccc} 
\hline\hline
\textbf{Problem} & \textbf{Method} & $C \downarrow$  & $\min(C)$ & $\max(C)$ & LVR[\%] $\downarrow$ & V[\%] & $\mathbb{E}(W_1) \uparrow$ & $N \uparrow$ & Time[min] $\downarrow$  \\ 
\hline
MBB              & ~               & ~                        & ~                  & ~                  & ~       & $53.5$           & $\times 10^{-2}$  & ~            & ~              \\
~                & FeniTop         & $0.68$                   & ~                  & ~                  & $0$     & $53.49$          & 0                 & $1$          & $2$            \\
~                & DB              & $0.67 \pm 0.01$          & $0.66$             & $0.67$             & $0$     & $53.49 \pm 0.00$ & $1.36$            & $2$          & $55.5$         \\
~                & TopoDiff        & $2.29 \pm 0.37$          & $1.79$             & $3.43$             & $0$     & $51.83 \pm 0.59$ & $2.14$            & ~            & ~              \\
~                & TOM             & $0.75 \pm 0.03$          & $0.66$             & $0.83$             & $0$     & $55.08 \pm 1.04$ & $4.23$            & ~            & $6.1$          \\ 
\hline
Cantilever       & ~               & ~                        & $\times 10^{-2}$   & ~                  & ~       & $50.00$          & $\times 10^{-2}$  & ~            & ~              \\
~                & FeniTop         & $0.59$                   & ~                  & ~                  & $0$     & $49.91$          & $0$               & $1$          & $4$            \\
~                & DB              & $0.58 \pm 0.01$          & $0.57$             & $0.58$             & $0$     & $49.98 \pm 0.00$ & $1.01$            & $2$          & $123$          \\
~                & TopoDiff*       & $1.19 \pm 0.28$          & $0.72$             & $1.95$             & $87$    & $50.23 \pm 0.70$ & $2.15$            & ~            & ~              \\
~                & TOM*            & $0.69 \pm 0.13$          & $0.64$             & $1.27$             & $8$     & $49.21 \pm 0.48$ & $2.42$            & ~            & $29$           \\ 
\hline
Bracket          & ~               & ~                        & $\times 10^{-3}$   & ~                  & ~       & $7.00$           & $\times 10^{-2}$  & ~            & ~              \\
~                & FeniTop         & $0.99$                   & ~                  & ~                  & $0$     & $7.16$           & $0$               & $1$          & $187$          \\
~                & TOM             & $1.31\pm0.01$            & $1.07$             & $1.54$             & $0$     & $6.86\pm0.09$    & $0.15$            & ~            & $47$          
\end{tabular}
    \label{tab:results}
    }
\end{table*}

\paragraph{TOM is more diverse and faster than DB.}
While the classical DB demonstrates higher compliance, TOM consistently outperforms DB in terms of diversity and wall-clock time, albeit with an increased number of solver steps. 
This performance advantage is primarily due to TOM's parallel nature, contrasting with the fundamentally sequential algorithm of DB. 
Additionally, the slower wall-clock time of DB can be partially attributed to its public implementation, which involves mesh refinement during training, thereby increasing the runtime for each solver step.

\paragraph{TOM learns a continuous space of shapes.}
Due to its modulated neural field architecture, TOM enables exploring different near-optimal solutions by moving continuously in the modulation space.

\paragraph{TOM has surface undulations.} 
Qualitatively, we observe, e.g., in Figure \ref{fig:results_simjeb}, that shapes produced by TOM are wavier than DB or TopoDiff and and can contain floaters (small disconnected components).
We attribute the undulations to the wavelet activation function of the WIRE architecture \citep{saragadam2023wire}.
Floaters are a common problem in neural TO approaches \citep{topodiff_maze22, nie2021topologygan} and can be corrected with simple post-processing \citep{subedi2020review}, which we demonstrate in Appendix \ref{app:postprocess}.

\paragraph{Ablations.}
We perform ablations to distill the impact of the diversity constraint, the radius of the modulation space, the SIMP penalty $p$, and annealing of the Heaviside filter.
We find that all these components are necessary for TOM to work.
Additionally, we show results for two alternative diversity metrics and demonstrate that they lead to similar conclusions.
Furthermore, we ablate the uniform sampling on the circle with fixed modulation vectors.
We find that this variant also leads to satisfactory results.
The details are provided in Appendix \ref{app:ablations}.

\section{Conclusion}

This paper presents topology optimization using modulated neural fields (TOM), a novel approach that addresses a limitation in traditional TO methods.
By leveraging neural networks to parameterize shapes and enforcing solution diversity as an explicit constraint, TOM enables the exploration of multiple near-optimal designs.
This is crucial for industrial applications, where manufacturing or aesthetic constraints often necessitate the selection of alternative designs.
Our empirical results demonstrate that TOM is both effective and scalable, generating more diverse solutions than prior methods while adhering to mechanical requirements.
Our main contributions include the introduction of TOM as the first data-free, solver-in-the-loop neural network training method and the empirical demonstration of its effectiveness across various problems.
This work opens new avenues for generative design approaches that do not rely on large datasets, addressing current limitations in engineering and design.

\paragraph{Limitations and future work.}
While TOM shows notable progress, several areas warrant further investigation to fully realize its potential.
An avenue for further investigation involves extending the one-dimensional modulation manifold to higher-dimensional spaces, thereby enabling exploration of designs across multiple axes within the modulation space.
Notably, the mitigation of floaters and undulations requires more in-depth investigation.
Additionally, more research is needed to explore different dissimilarity metrics for the diversity loss, as this could further enhance the variety of generated designs.
Improving sample efficiency is also crucial for practical applications, as the current method may require significant computational resources.
Promising directions include using second-order optimizers \citep{jnini2024gauss}. 
Lastly, future work could train TOM on coarse designs and subsequently refine them with classical TO methods, combining the strengths of both approaches to achieve even better results.

\begin{ack}

The ELLIS Unit Linz, the LIT AI Lab, the Institute for Machine Learning, are supported by the Federal State Upper Austria. We thank the projects FWF AIRI FG 9-N (10.55776/FG9), AI4GreenHeatingGrids (FFG- 899943), Stars4Waters (HORIZON-CL6-2021-CLIMATE-01-01), FWF Bilateral Artificial Intelligence (10.55776/COE12). We thank the European High Performance Computing initiative for providing computational resources (EHPC-DEV-2023D08-019, 2024D06-055, 2024D08-061). We thank NXAI GmbH, Audi AG, Silicon Austria Labs (SAL), Merck Healthcare KGaA, GLS (Univ. Waterloo), T\"{U}V Holding GmbH, Software Competence Center Hagenberg GmbH, dSPACE GmbH, TRUMPF SE + Co. KG.

\end{ack}

\medskip

\small

\bibliographystyle{styles/icml25/icml2025}
\bibliography{99_References}

\newpage
\appendix
\setcounter{equation}{0}
\renewcommand{\theequation}{\thesection\arabic{equation}}

\newpage

\section{Implementation and experimental details}
\label{app:exp_details}

\subsection{Algorithmic description}

The TOM algorithm is given in Algorithm \ref{alg:TOM}. A detailed textual description is provided in Section~\ref{sec:method}.

\begin{algorithm}[b]
   \caption{TOM}
   \label{alg:TOM}
\begin{algorithmic}
   \STATE {\bfseries Input:} 
   parameterized density $f_\theta$, 
   vector of mesh points $\mathbf{x_i} \in \mathbb{R}^{d_x}$,
   $\beta$ and annealing factor $\Delta_\beta$,
   number of shapes per iteration $k$,
   learning rate scheduler $\gamma(t)$,
   iterations $T$
   \FOR{$t=1$ {\bfseries to} $T$}
   \STATE $\mathbf{z_j} \sim p(\mathcal{Z})$ \hfill\COMMENT{sample modulation vectors}
   \STATE $\Tilde{\rho}_{z_j} \leftarrow f_\theta(\mathbf{x_i}, \mathbf{z_j})$  \hfill\COMMENT{net forward all shapes $j$}
   \STATE $\rho_j \leftarrow H(\Tilde{\rho}_{z_j}, \beta)$ \hfill\COMMENT{Heaviside contrast filter}
   \STATE $C, V, \frac{\partial C}{\partial \rho}, \frac{\partial V}{\partial \rho} \leftarrow \text{FEM}(\rho_j)$ \hfill\COMMENT{solver step}
   \STATE $\frac{\partial \rho}{\partial \theta} \leftarrow \text{backward}(f_\theta, \rho_j)$ \hfill\COMMENT{autodiff backward}
   \STATE $\nabla_\theta C \leftarrow \frac{\partial C}{\partial \rho} \frac{\partial \rho}{\partial \theta} $ \hfill\COMMENT{compliance gradient}
    \STATE $\nabla_\theta V \leftarrow \lambda_V \frac{\partial C}{\partial \rho} \frac{\partial \rho}{\partial \theta} $ \hfill\COMMENT{volume gradient}
   \STATE $L_\text{G} \leftarrow \lambda_\text{interface} L_\text{interface} + ...$ \ \hfill\COMMENT{GINN constraints}
   \STATE $\Delta \theta \leftarrow  \text{ALM} \left( \nabla_\theta C, \nabla_\theta V, \nabla_\theta L_\text{G} \right) $ \hfill\COMMENT{ALM}
   \STATE $\theta \leftarrow \theta - \gamma(t) \Delta \theta $ \hfill\COMMENT{parameter update}
   \STATE $\beta \leftarrow \beta \Delta_\beta$ \hfill\COMMENT{annealing}
   \ENDFOR
\end{algorithmic}
\end{algorithm}

\subsection{Additional results}
\label{app:additional}

Table \ref{tab:additional_results} contains additional characterization of the main experimental results in terms of computational resources.

\begin{table}
    \centering
    \caption{
       Additional characterization of the main experimental results reported in Table \ref{tab:results}, including iterations needed to converge.
       Each iteration consists of several FEM solver calls, which are executed in parallel for TOM, resulting in an overall faster training than the sequential DB.
       Lastly, we report the mesh resolution used by the FEM solver. 
       For DB, it changes over training due to the adaptive mesh refinement, which is indicated by the initial grid $\rightarrow$ final refined mesh size. 
    }
    \label{tab:additional_results}
    \begin{tabular}{llccc} 
    \hline\hline
    Problem & Method & Iterations & Total solver calls & FEM resolution     \\ 
    \hline
    MBB              & ~               & ~             & ~              & ~                           \\
    ~                & FeniTop         & $400$         & $400$          & $180 \times 60$                      \\
    ~                & DB              & ~             & $3190$         & $50 \times 150 \rightarrow 27174$  \\
    ~                & TopoDiff        & ~             & ~              & $64 \times 64$                     \\
    ~                & TOM             & $400$         & $10000$        & $180 \times 60$                      \\ 
    \hline
    Cantilever       & ~               & ~             & ~              & ~                           \\
    ~                & FeniTop         & $400$         & $400$          & $150 \times 100$                     \\
    ~                & DB              & ~             & $4922$         & $25 \times 37 \rightarrow 52024$   \\
    ~                & TopoDiff*       & ~             & ~              & $64 \times 64$                       \\
    ~                & TOM*            & $1000$        & $25000$        & $150 \times 100$                     \\ 
    \hline
    Bracket          & ~               & ~             & ~              & ~                           \\
    ~                & FeniTop         & $400$         & $400$          & $104 \times 172 \times 60$                  \\
    ~                & TOM             & $1000$        & $9000$         & $26 \times 43 \times 15$                   
    \end{tabular}
\end{table}

\subsection{Model hyperparameters}

The most important hyperparameters are summarized in Table~\ref{tab:hps}.

\paragraph{Compute.} 
We report additional information on the experiments and their implementation.
We run all experiments on a single GPU (NVIDIA Titan V12) on a node with a Xeon Gold 6150 CPU (36 Cores, 2.70GHz) and 384GB RAM.
The maximum GPU memory requirements are less than 2GB for all experiments.

\paragraph{Neural network.} 
For the model to effectively learn high-frequency features, it is important to use a neural network represenation with a high frequency bias \citep{sitzmann2019siren,Teney2024NeuralRR}.
Hence, all models were trained using the real, 1D variant of the WIRE architecture \citep{saragadam2023wire}.
WIRE allows to adjust the frequency bias by setting the hyperparameters $\omega_0^1$ and $s_0$.
For this architecture, each layer consists of 2 Multi-layer perceptron (MLPs), one has a periodic activation function $cos(\omega_0 x)$, the other with a gaussian $e^{(s_0 x)^2}$. 
The post-activations are then multiplied element-wise.

\paragraph{Optimization.}
The overall loss for a data point \( x \) can be expressed as \( L(x) = o(x) + \sum_i \lambda_i c_i(x) \), where \( o(\cdot) \) represents the objective function, \( c_i(\cdot) \) denotes the constraints, and \( \lambda_i \) are the balancing coefficients for these constraints.
We employ the augmented Lagrangian method (ALM) \citep{basir_adaptive_2023} to dynamically balance the various constraints throughout the training process. Intuitively, the better a constraint \( c_i \) is satisfied, the smaller the corresponding \( \lambda_i \).
We designate the compliance loss as the objective function, which consequently remains unaffected by the balancing mechanism of ALM.
Crucially, for ALM to function optimally, we scale all loss terms by a dedicated factor to ensure they are approximately on the same magnitude. 
This adjustment is analogous to harmonizing the different ``units'' of the losses (e.g., volume loss versus diversity loss).

\paragraph{Delayed start of diversity loss for jet engine bracket.}
For the jet engine bracket, we start the chamfer discrepancy loss at an iteration where the shapes already have reasonable formed (see ``start diversity'' in Table \ref{tab:hps}).
Empirically, this is needed to not destroy the shapes early on.
This could have several potential reasons, including a low target volume of only 0.07 or the fact that 3D shape optimization might be fundamentally harder.
We leave further investigations to future work.

\begin{table}[!ht]
    \caption{TOM hyperparameters for different experiments.}
    \centering
    \begin{tabular}{lccc}
        \toprule
        & \textbf{MBB beam} & \textbf{Cantilever} & \textbf{JEB} \\ 
        \hline
        Hidden layers & 32x3 & 32x3 & 64x3 \\ 
        $\omega_0^1$ for WIRE & 10 & 9 & 18 \\ 
        $s_0$ for WIRE 10 & 10 & 6  \\ 
        Learning rate & $5 \cdot 10^{-5}$  & $5 \cdot 10^{-5}$ & $10^{-3}$ \\ 
        Decay rate & 400  & 200 & ~ \\ 
        \hline
        radius $r$ & 1.2 & 0.6 & 0.02 \\ 
        SIMP penalty $p$ & 3 & 3 & 1.5 \\ 
        $\beta$ annealing & $[0, 400]$ & $[0, 400]$ & $[0, 400]$ \\ 
        minimal diversity $\delta^* $ & 0.3 & 0.4 & 0.13 \\ 
        \# iterations & 400 & 1,000 & 1,000 \\ 
        \# shapes per batch & 25 & 25 & 9 \\ 
        \hline
        Compliance scale & 1 & 1 & 1,000 \\ 
        Volume scale & 1 & 1 & 10 \\ 
        Chamfer diversity scale & 1 & 10 & 10 \\ 
        Interface scale & ~ & ~ & 2,000 \\ 
        Interface normal scale & ~ & ~ & 1 \\ 
        Design region scale & ~ & ~ & 1 \\ 
        \hline
        Start diversity & ~ & ~ & 350 \\ 
        \bottomrule
    \end{tabular}
    \label{tab:hps}
\end{table}

\subsection{Reference solutions} \label{subsec:ref_solutions}
\begin{figure*}[ht!]
\centering
\begin{subfigure}[t]{0.3\textwidth}
    \centering
    \includegraphics[width=\textwidth]{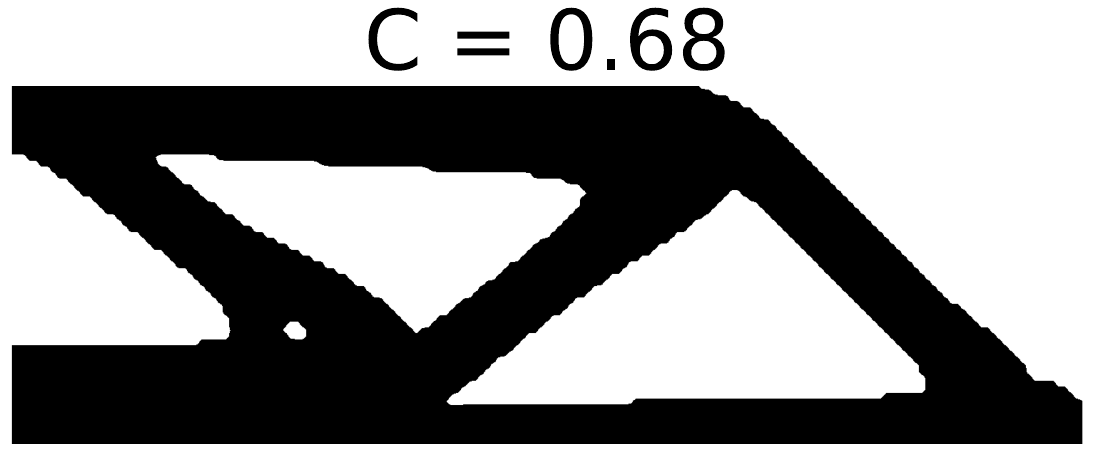}
    \caption{Single solution generated with FeniTop}%
    \label{fig:fenitop_mbb}
\end{subfigure}
\begin{subfigure}[t]{0.3\textwidth}
    \centering
    \includegraphics[width=\textwidth]{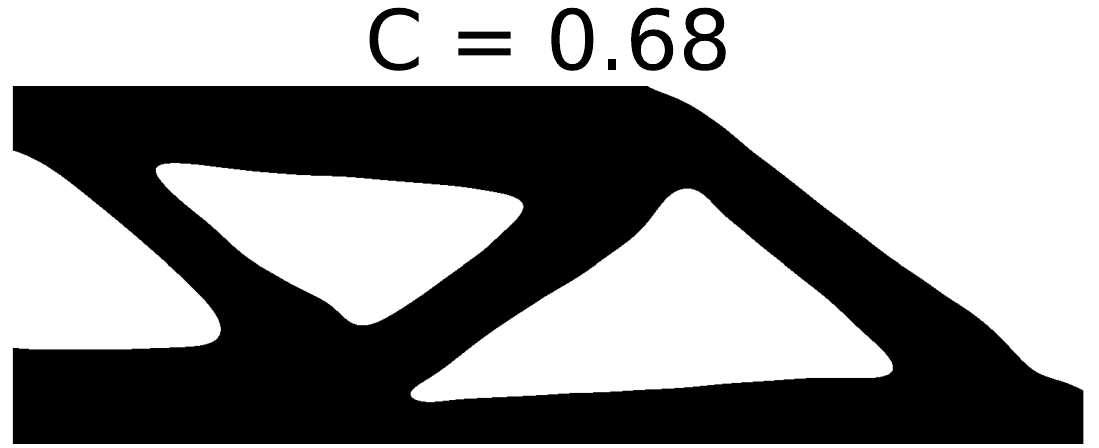}
    \caption{Single solution generated with TOM}%
    \label{fig:TOM_single_mbb}
\end{subfigure}

\caption{ Reference solutions for the MBB beam problem
}
\label{fig:mbb_reference}
\end{figure*}

\begin{figure*}[ht!]
\centering
\begin{subfigure}[t]{0.25\textwidth}
    \centering
    \includegraphics[width=\textwidth]{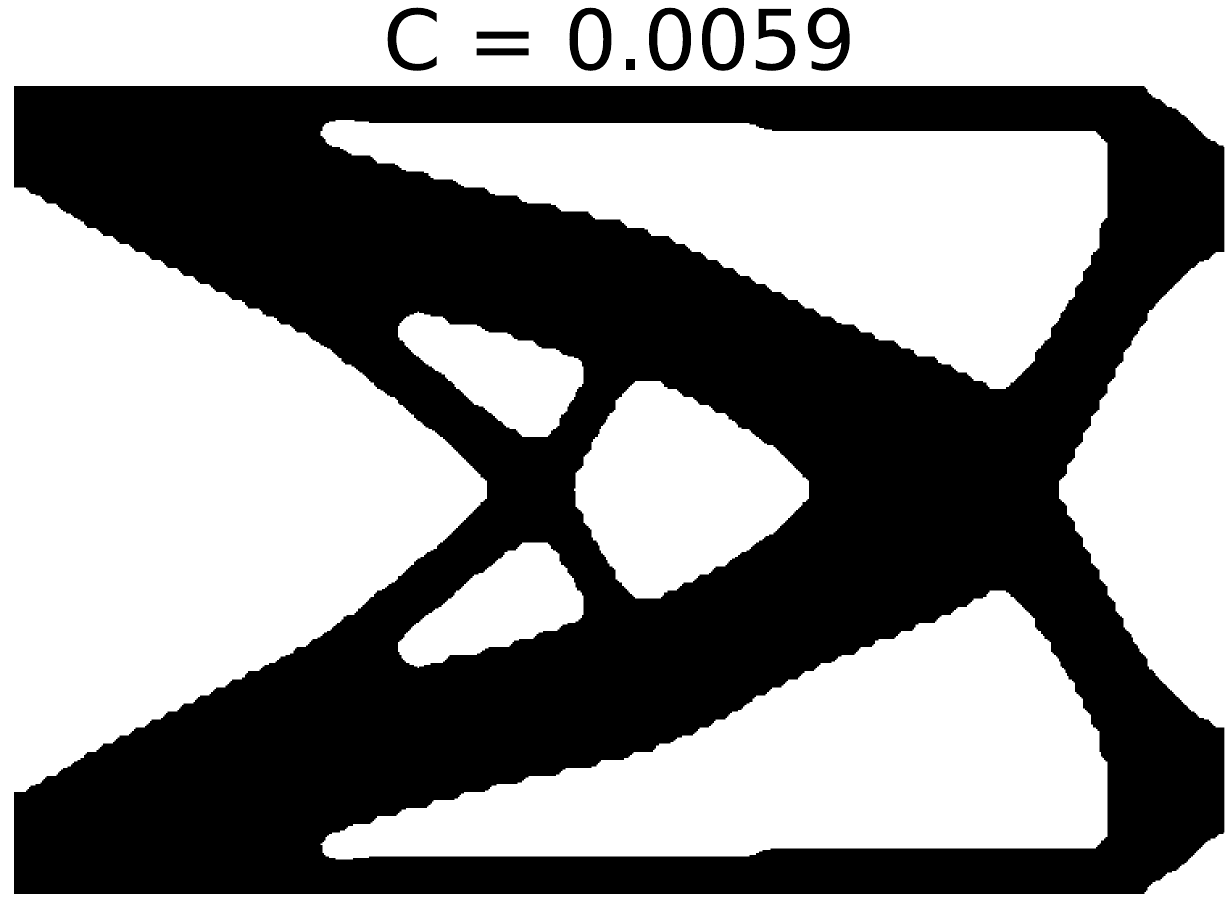}
    \caption{Single solution generated with FeniTop}%
    \label{fig:fenitop_cantilever}
\end{subfigure}
\begin{subfigure}[t]{0.25\textwidth}
    \centering
    \includegraphics[width=\textwidth]{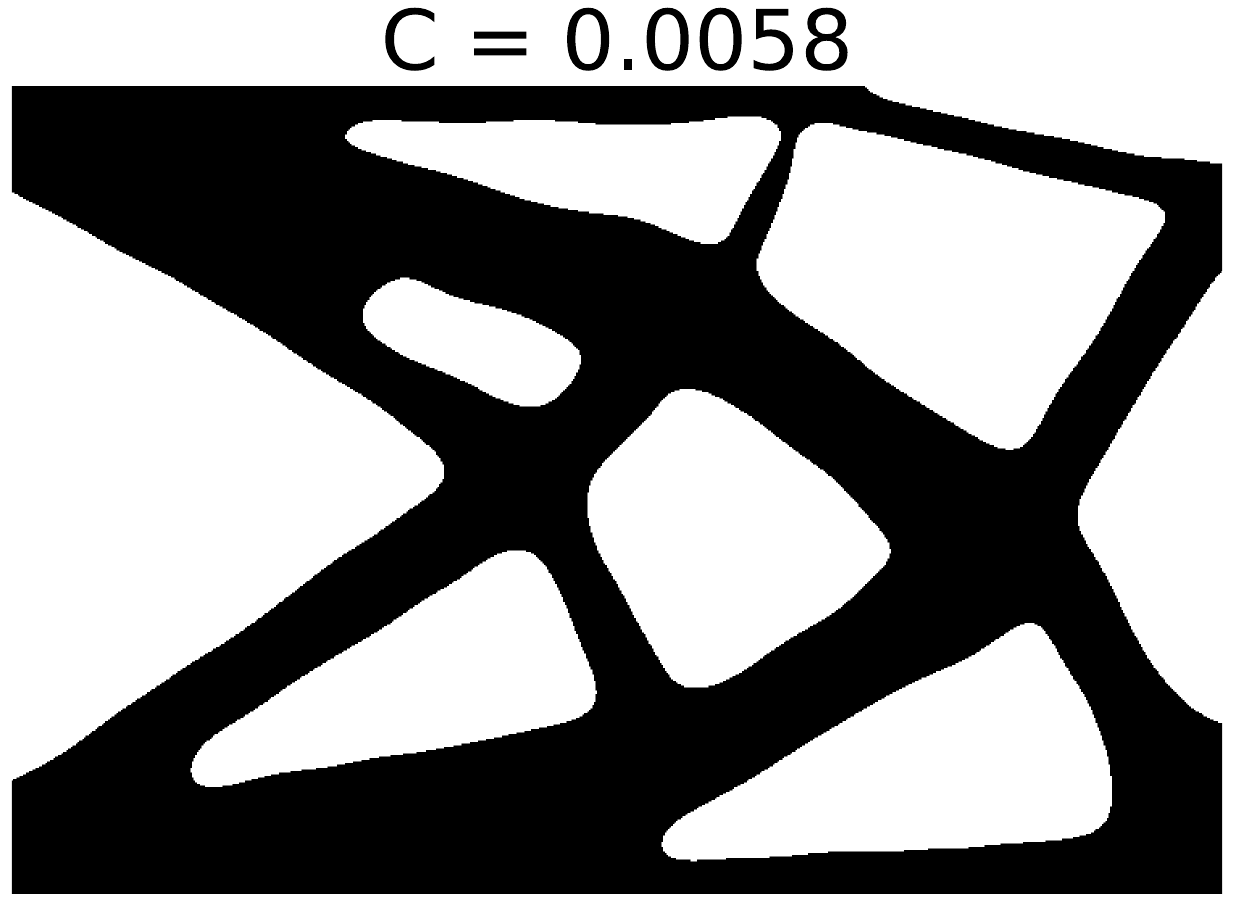}
    \caption{Single solution generated with TOM}%
    \label{fig:TOM_single_cantilever}
\end{subfigure}

\caption{Reference solutions for the Cantilever problem}
\label{fig:cantilever_reference}
\end{figure*}

\begin{figure*}[ht!]
\centering
\begin{subfigure}[t]{0.4\textwidth}
    \includegraphics[width=\textwidth]{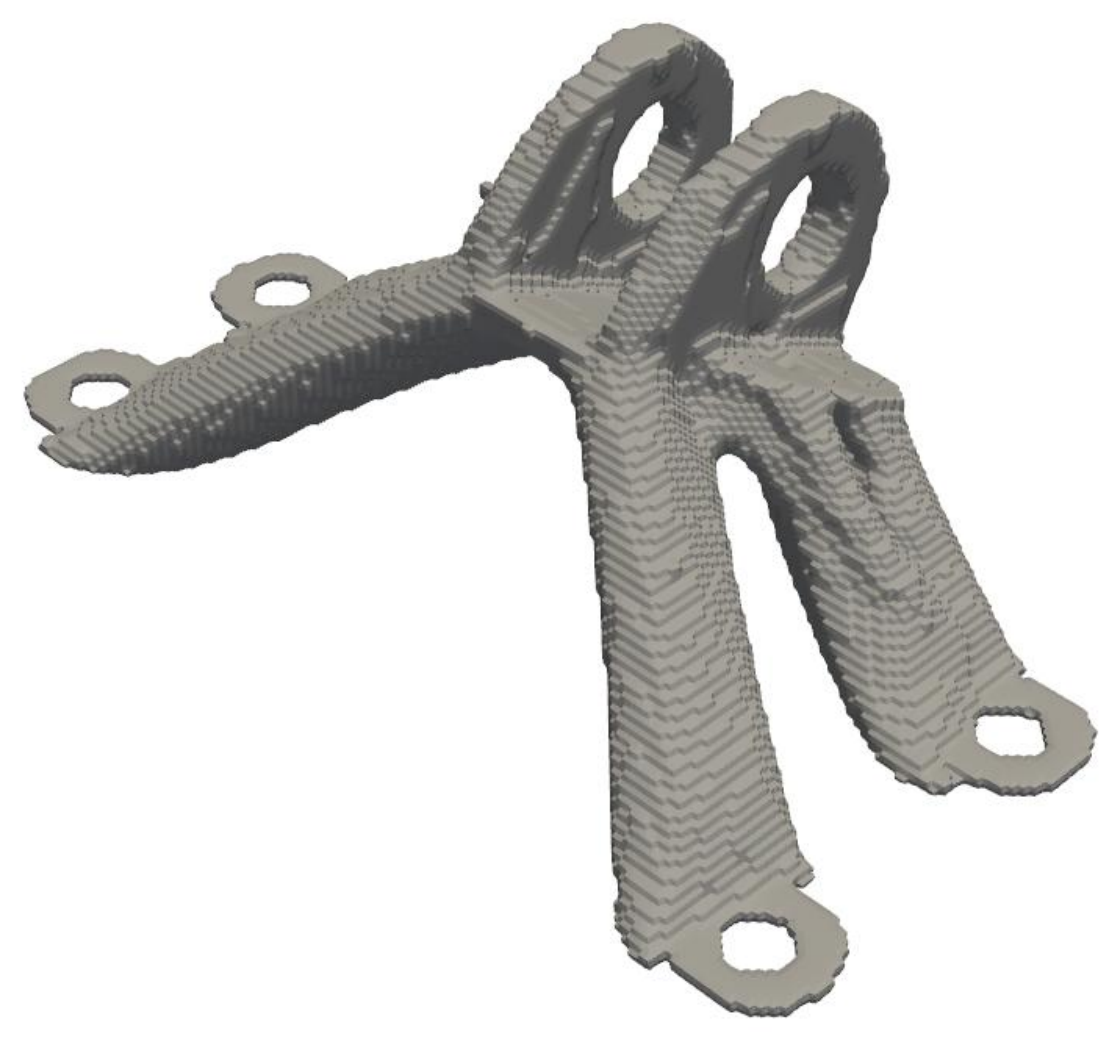}
    \caption{Single solution generated with FeniTop}
    \label{fig:simjeb_fenitop}   
\end{subfigure}
\begin{subfigure}[t]{0.4\textwidth}
    \includegraphics[width=\textwidth]{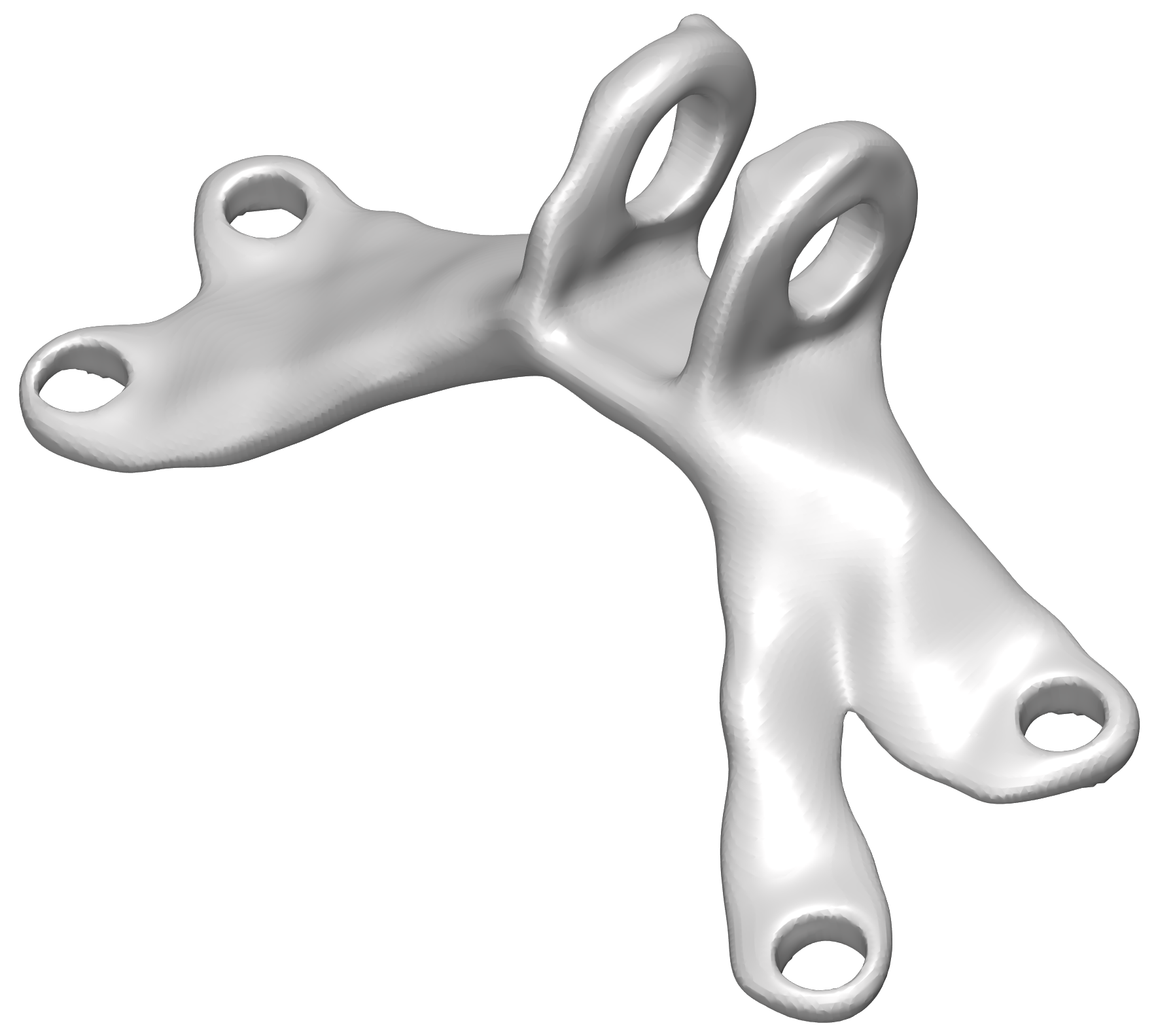}
    \caption{Single solution generated with TOM}
    \label{fig:TOM_single_simjeb}   
\end{subfigure}
\caption{Reference solutions for the jet engine bracket (JEB) problem}
\label{fig:simjeb_reference}
\end{figure*}

We provide reference solutions to the problem settings generated by the standard TO. We use the implementation of the SIMP method provided by the FeniTop library \cite{fenitop} as the classical TO baseline. \newline
We also generate single solutions using TOM without a diversity constraint or modulation variable as input. These single shape training runs showcase the baseline capability of TOM. \newline
The reference solutions for MBB beam problem are shown in Figure \ref{fig:mbb_reference}, for the cantilever beam problem in Figure \ref{fig:cantilever_reference}, and for the jet engine bracket problem in Figure \ref{fig:simjeb_reference}.

\FloatBarrier

\subsection{Load violations in TopoDiff}
\label{app:LV}

\cite{topodiff_maze22} provide a detailed evaluation of TopoDiff, including a test data set (called \textit{level 2 test data}) with unseen boundary conditions. They report that TopoDiff successfully generalizes to unseen boundary conditions and produces shapes which fit these new, unseen boundaries with $100\%$ accuracy ($0\%$ load violations, Table 1 in \cite{topodiff_maze22}). 

\begin{figure}[htb]
    \centering
    \begin{subfigure}[T]{0.69\textwidth}
    \centering %
        \includegraphics[height=0.25\textheight, keepaspectratio]{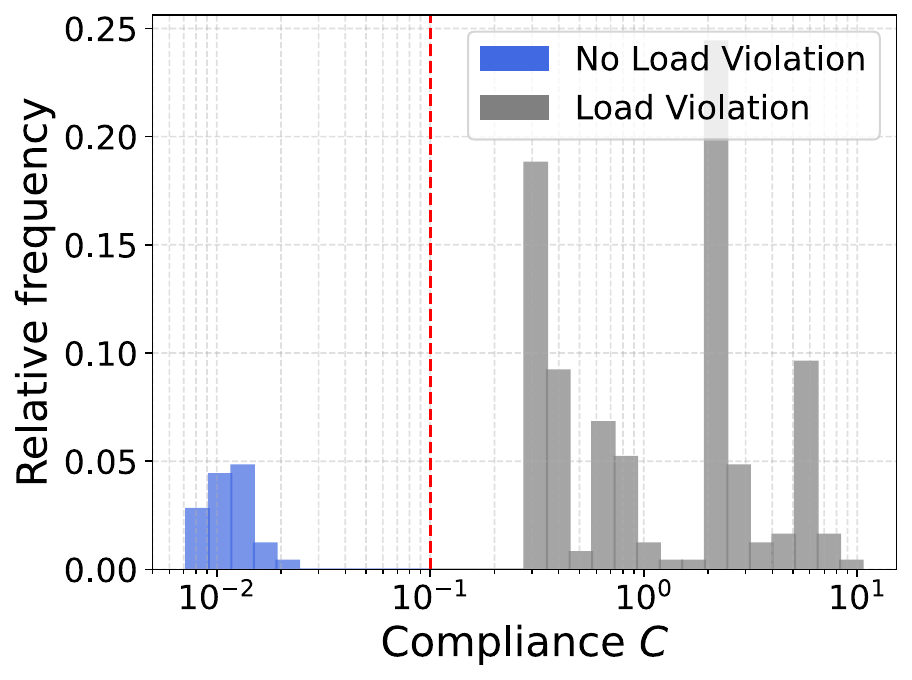}
        \caption{Distribution of compliance $C$ of TopoDiff shapes. The large jump in compliance above $10^{-1}$ is explained by forces being applied to the ersatz material.}
        \label{fig:c_histo_topodiff}
    \end{subfigure}
    \hfill %
    \begin{subfigure}[T]{0.3\textwidth}
        \centering %
        \includegraphics[height=0.25\textheight, keepaspectratio]{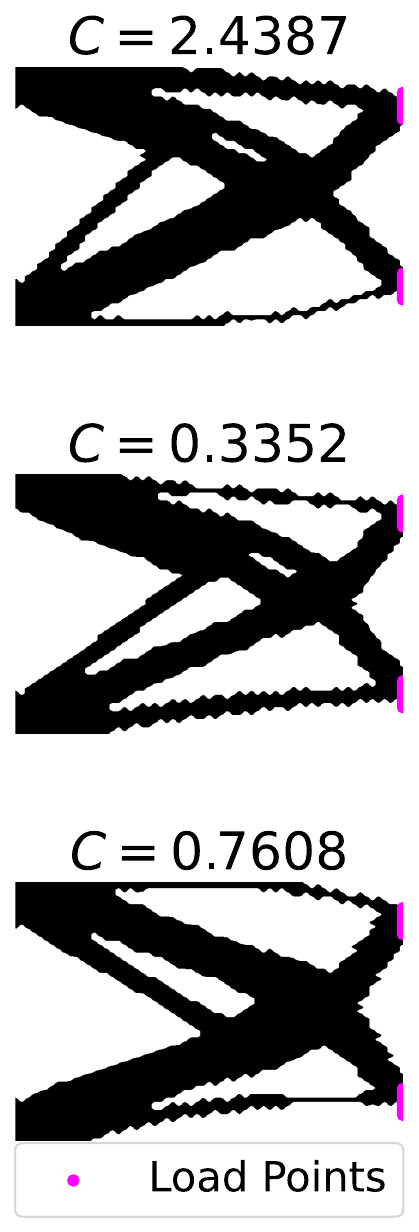}
        \caption{Load violations of TopoDiff shapes. Shapes do not attach to all load points, leading to artifacts in the computed compliance $C$.}
        \label{fig:load_violation_topodiff}
    \end{subfigure}
\end{figure}

In the case of the 2D cantilever problem, we find that $87.2\%$ of the shapes generated by TopoDiff contain load violations. Our FEM solver regularizes the solve step by placing so-called ersatz material with a very low (but non-zero) density at void mesh elements. This prevents NaNs or undefined behavior by the solver when encountering violated boundary conditions, such as no material at loading points (load violations), see Figure \ref{fig:load_violation_topodiff}.

This results in a large jump in the compliance values, which can be seen in Figure \ref{fig:c_histo_topodiff}. Note that these values are, however, unphysical and an artifact of the solver using ersatz-material instead of void cells. Physically, the compliance is simply undefined as a force cannot be applied when no material is present.

We filter out all shapes containing load violations in the results section to allow for a better comparison and to avoid reporting compliances that are mere solver artifacts and not physically meaningful.

\subsection{Ablations}
\label{app:ablations}

To better understand the impact of different parts of TOM, we perform ablations and summarize the results in Table \ref{tab:ablation_diversity}.

\begin{table}[h]
\caption{Results for ablations on the three TO problems.}
\centering
\LARGE
\resizebox{\textwidth}{!}{

\begin{tabular}{llcccccccc}
\hline
\hline
Problem & Method & $C$ & $\min(C)$ & $\max(C)$ & V[\%] & LVR[\%] & $\mathbb{E}(W1)$ & $\mathbb{E}(d_H)$ & $\mathbb{E}(d_\text{SSIM})$ \\
\hline
MBB Beam &  &  &  &  &  &  &  \\
 & Base & $0.75 \pm 0.03$ & $0.66$ &$ 0.83$ & $55.08 \pm 1.04$ & $0.0$ & $0.042$ & $186.22$ & $0.1096$\\
 & No diversity constraint & $0.76 \pm 0.02$ &$ 0.73$ & $0.84$ & $54.62 \pm 1.07$ & $0.0$ & $0.042$ & $170.10$ & $0.1124$\\
 & Radius 1/10 & $0.85 \pm 0.23$ & $0.65$ & $1.43$ & $58.55 \pm 2.63$ & $0.0 $& $0.040$ & $93.29$ & $0.1075$ \\
 & Penalty $p=1.5$ & $1.56 \pm 1.31$ & $0.68$ & $5.54$ & $58.17 \pm 2.93$ & $0.0$ & $0.043$ & $169.28$ & $0.1299$\\
 & Annealing & $ 0.89 \pm 0.25 $ & $ 0.67 $ & $ 1.66 $ & $ 59.75 \pm 1.70 $ & $ 0.0 $ & $ 0.028 $& $221.58$ & $0.0917$\\
 & Fixed modulation & $0.71 \pm 0.01$ & $0.70$ & $0.76$ &$ 53.49 \pm 0.32$ & $0.0$ & $0.022$ & $197.55$ & $0.0784$\\
\hline
Cantilever &  &  &  &  &  &  &  \\
 & Base  & $ 0.69 \pm 0.13 $ & $ 0.64 $ & $ 1.28 $ & $ 49.21 \pm 0.48 $ & $ 8.0 $ & $ 0.024 $ & $85.33$ & $0.1799$\\
 & No diversity constraint & $ 0.90 \pm 0.88 $ & $ 0.62 $ & $ 4.95 $ & $ 49.63 \pm 0.61 $ & $ 8.0 $ & $ 0.023 $ & $89.25$ & $0.1791$\\
 & Radius 1/10 & $ 0.92 \pm 0.07 $ & $ 0.79 $ & $ 1.01 $ & $ 49.43 \pm 2.08 $ & $ 0.0 $ & $ 0.032 $ & $84.02$ & $0.1186$\\
 & Penalty $p=1.5$ & $ 1.19 \pm 1.21 $ & $ 0.54 $ & $ 4.97 $ & $ 50.55 \pm 1.15 $ & $ 12.0 $ & $ 0.019 $ & $69.14$ & $0.1565$\\
 & Annealing  & $ 0.74 \pm 0.22 $ & $ 0.61 $ & $ 1.35 $ & $ 49.53 \pm 0.33 $ & $ 12.0 $ & $ 0.027$  & $84.42$ & $0.1767$\\
 & Fixed modulation  & $ 0.82 \pm 0.26 $ & $ 0.62 $ & $ 1.25 $ & $ 48.87 \pm 0.28 $ & $ 4.0 $ & $ 0.026$  & $93.93$ & $0.1848$\\
\hline
Simjeb &  &  &  &  &  &  &  \\
 & Base & $ 0.13 \pm 0.01 $ & $ 0.11 $ & $ 0.15 $ & $ 6.86 \pm 0.09 $ & $ 0.0 $ & $ 0.027 $ & $9.01$ & $0.0782$\\
 & No diversity constraint  & $ 0.11 \pm 0.01 $ & $ 0.10 $ & $ 0.14 $ & $ 7.51 \pm 0.06 $ & $ 0.0 $ & $ 0.006 $ & $8.2$ & $0.0687$\\
 & Radius 1/10 & $ 0.13 \pm 0.01 $ & $ 0.12 $ & $ 0.15 $ & $ 7.35 \pm 0.02 $ & $ 0.0 $ & $ 0.002$  & $5.54$ & $0.0093$\\
 & Penalty $p=1.5$ & $ 0.09 \pm 0.00 $ & $ 0.09 $ & $ 0.10 $ & $ 7.09 \pm 0.01 $ & $ 0.0 $ & $ 0.010$  & $11.33$ & $0.0443$\\
 & Annealing  & $ 0.60 \pm 0.19 $ & $ 0.40 $ & $ 0.85 $ & $ 6.21 \pm 0.05 $ & $ 66.7 $ & $ 0.041$ & $9.47$ & $0.1024$\\
 & Fixed modulation & $ 0.08 \pm 0.01 $ & $ 0.07 $ & $ 0.09 $ & $ 7.64 \pm 0.04 $ & $ 0.0 $ & $ 0.034$ & $11.01$ & $0.0970$
\end{tabular}
\label{tab:ablation_diversity}
}
\end{table}

\begin{figure}
    \centering
    \includegraphics[width=0.5\linewidth]{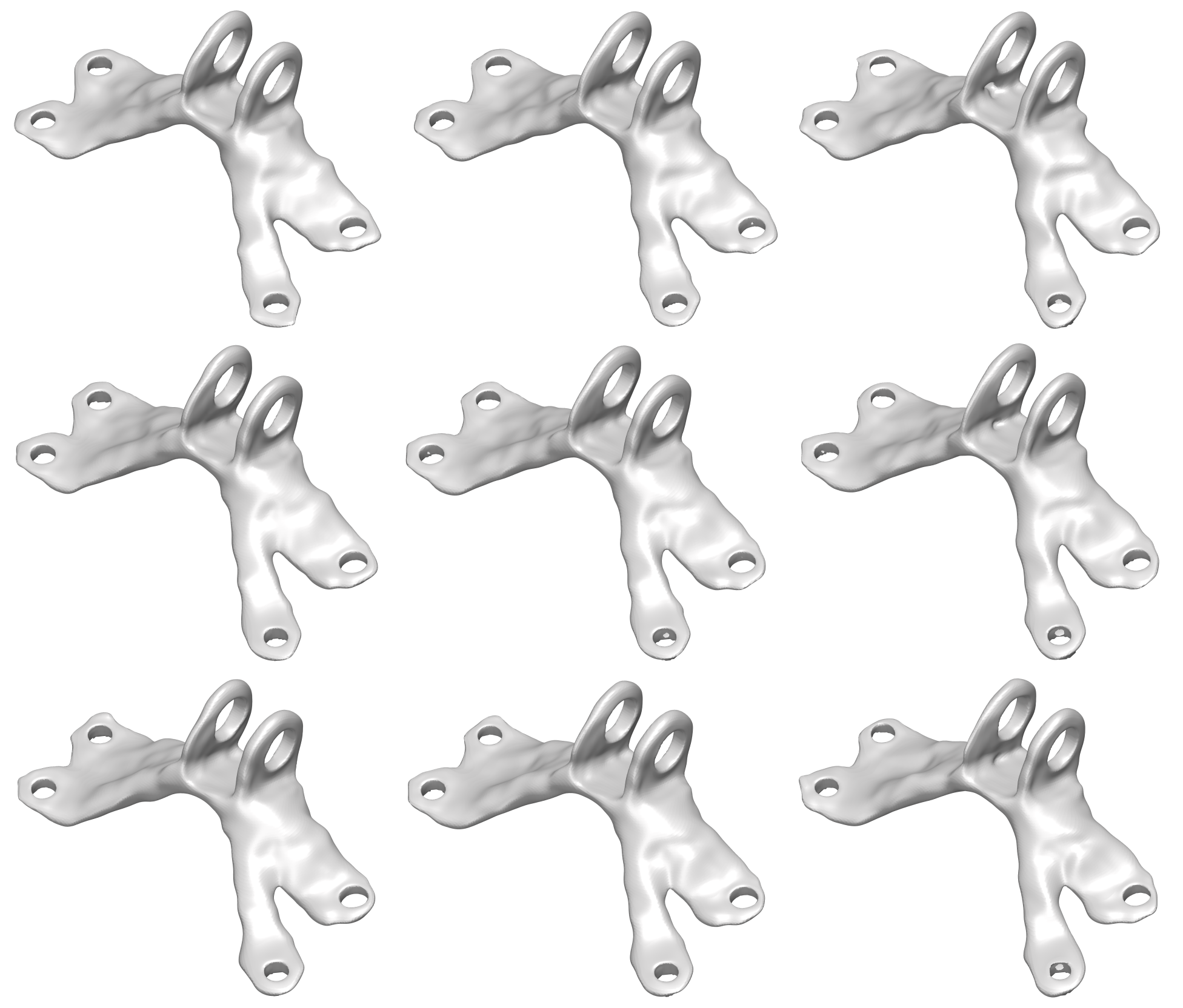}
    \caption{JEB: Training in 3D without diversity constraint results in mode collapse.}
    \label{fig:diversity_ablation_jeb}
\end{figure}

\paragraph{Diversity constraint.} 
We ablate the diversity loss term. 
While some diversity remains in the 2D case due to the network initialization randomness, we observe complete mode collapse in 3D, illustrated in Figure \ref{fig:diversity_ablation_jeb}.

\paragraph{Radius $r$ of modulation space.}
We highlight the choice of radius $r$ by ablating it with a radius $r' = \frac{1}{10} r$.
The resulting shapes are less diverse (c.f. Table \ref{tab:ablation_diversity}) despite applying the diversity constraint during training.
\\
The importance of a network's frequency bias has been highlighted by several works \citep{sitzmann2019siren,saragadam2023wire}.
Importantly, depending on the problem, there might be a different frequency bias necessary for the coordinates $\mathbf{x}$ and the modulation vectors $\mathbf{z}$.
For the WIRE architecture we use in our experiments, the frequency bias of the modulation is implicitly controlled by the radius $r$.
The greater $r$, the greater the diversity at initialization and convergence as suggested by \citet{Teney2024NeuralRR}.

\begin{figure*}[ht!]
\centering
\begin{subfigure}[t]{0.4\textwidth}
    \centering
    \includegraphics[width=\columnwidth]{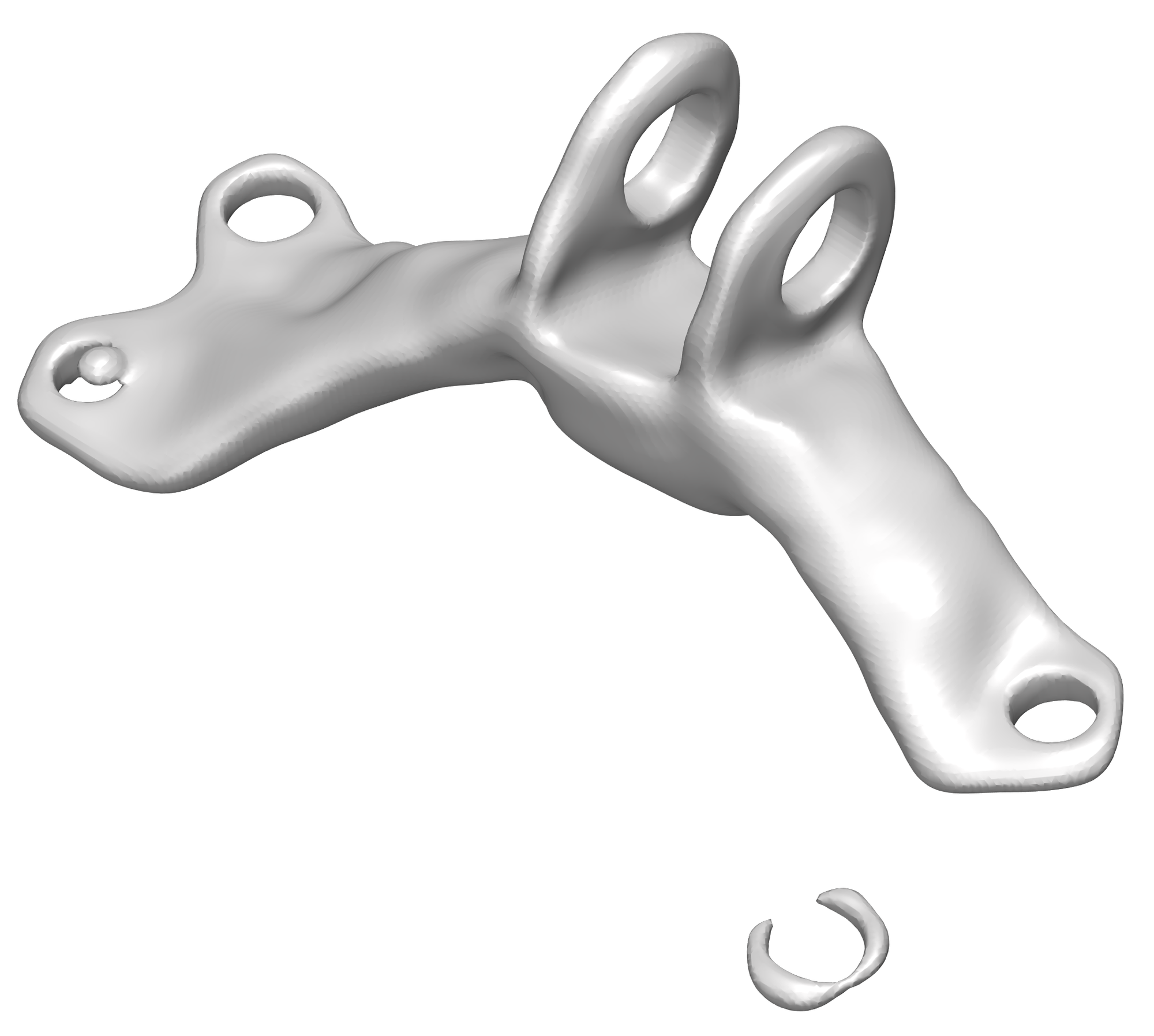}
    \caption{}%
    \label{fig:jeb_p3}
\end{subfigure}
\hfill
\begin{subfigure}[t]{0.4\textwidth}
    \centering
    \includegraphics[width=\columnwidth]{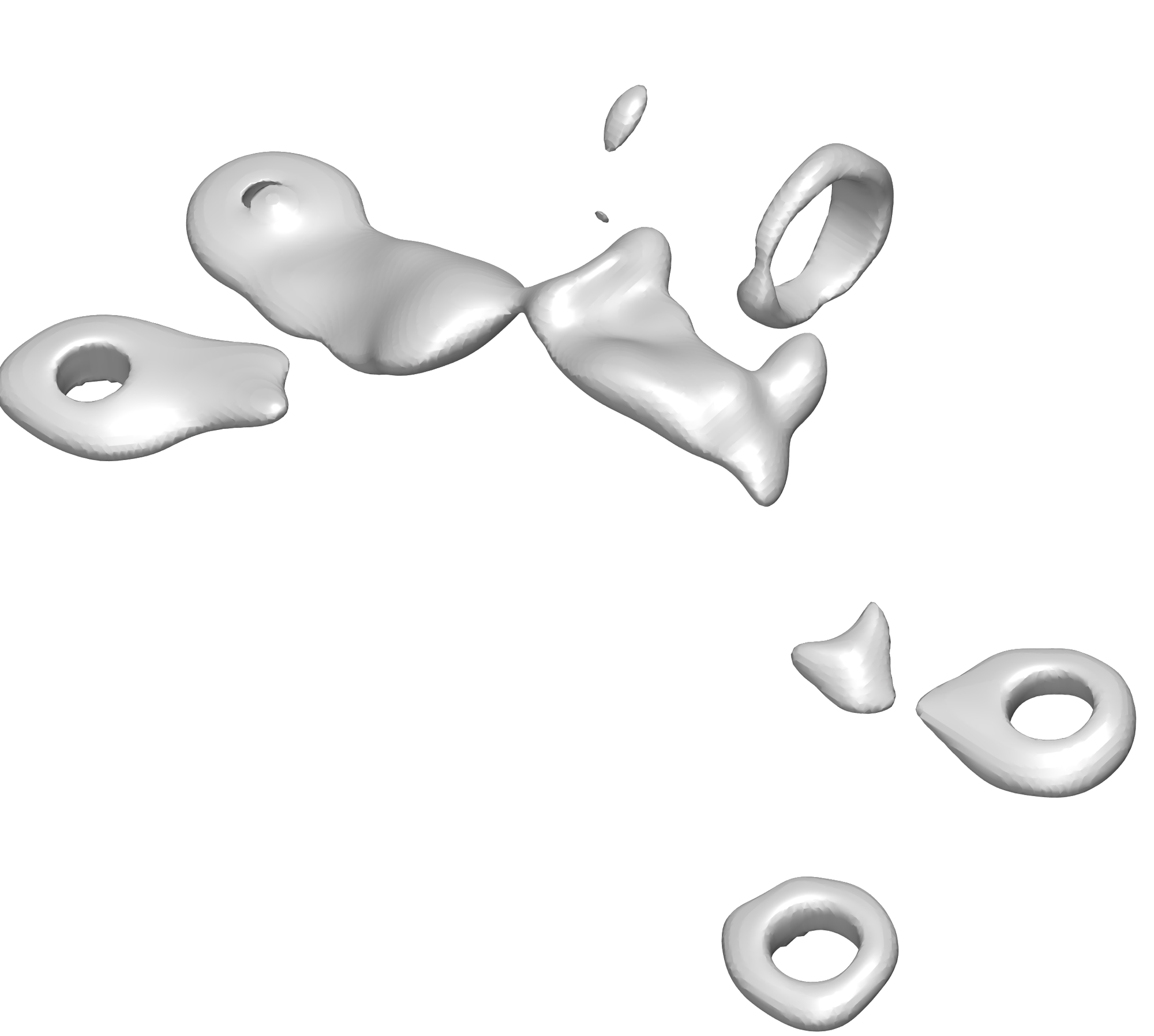}
    \caption{}%
    \label{fig:jeb_no_annealing}
\end{subfigure}

\caption{
Ablations for penalty and $\beta$ annealing.
(a) Jet engine bracket trained with penalty $p=3$.
(b) Jet engine bracket trained without annealing.
}
\label{fig:ablations_annealing_penalty}
\end{figure*}

\paragraph{Sensitivity to penalty $p$.}
As noted by prior work \citep{sanu2024neural_good_bad_ugly}, neural reparemeterization is sensitive to the penalty parameter $p$.
For SIMP, the density is penalized with an exponent $p$ (see Section \ref{subsec:TO}).
We ablate this parameter by setting the 2D penalty from $p=3$ to $1.5$ and for 3D from $p=1.5$ to $3$.
Our experiments confirm that TOM is sensitive to the choice of $p$, as it sharpens the loss landscape.
This becomes more apparent when looking at the derivative of the stiffness of a mesh element $\frac{\partial \mathbf{K}_i(\rho_i)}{\partial \rho_i} = p \rho_i^{p-1} \mathbf{K}_i$.
E.g., for $p=3$ the gradient is quadratically scaled by the current density value. 
This implies that the higher $p$, the harder it is to escape local minima.
We do not observe a large performance impact in the 2D experiments. However, in 3D, we observe that $p=3$ instead of $p=1.5$ leads to convergence to an undesired local minimum, illustrated in Figure \ref{fig:jeb_p3}.

\paragraph{Annealing is necessary for good convergence.}
We find that TOM requires annealing to achieve good convergence.
Figure \ref{fig:jeb_no_annealing} demonstrates a failed run without $\beta$ scheduling the Heaviside function (Equation \ref{eq:heaviside}).
The optimization fails to converge to a useful compliance value and does not fulfill the desired interface constraints.

\paragraph{Fixed modulation vectors.}
TOM uses a continuous modulation space, and in our current implementation, samples modulation vectors from a circle.
An interesting alternative is to keep the modulation vectors fixed throughout training. 
As shown in Table \ref{tab:ablation_diversity}, the results of this variant also produce shapes with satisfactory compliance and diversity.

\begin{figure}
    \centering
    \includegraphics[width=0.5\linewidth]{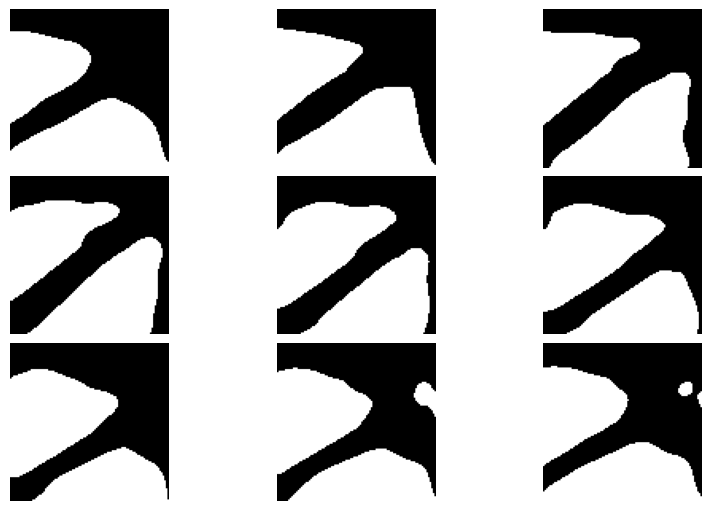}
    \caption{The cantilever ablation of the radius leads to mode collapse qualitatively.}
    \label{fig:ablation_cantilever_radius}
\end{figure}

\paragraph{Alternative diversity metrics}
We measure diversity via the Hill number $D_q(S)$ over the set $S$, where we choose $q=2$ as it is interpretable as the expected dissimilarity between two elements drawn from $S$.
The diversity can be computed based on an arbitrary pairwise dissimilarity function.
As a measure of dissimilarity, we chose the Wasserstein distance because we consider material distributions, and the Wasserstein distance is a distance between distributions. Also, it is a proper metric distance satisfying the four axiomatic properties.
\\
In addition to the Wasserstein-1 distance, we performed additional evaluations of the diversity based on two other dissimilarity functions: the Hausdorff distance ($d_H$) and structural dissimilarity ($d_\text{SSIM}$) \citep{wang2004image}. The results in Table \ref{tab:ablation_diversity} show similar results across all dissimilarity functions. 
Note that the Hausdorff distance is particularly sensitive to outliers, making it less suitable for robust measurements.
\\
Additionally, we observe that the proposed diversity metrics do not fully capture our intuition about the variety of shapes. 
For instance, the radius-ablation variant of the cantilever exhibits a higher diversity score than the base model. 
However, the qualitative evaluation in Figure \ref{fig:ablation_cantilever_radius} reveals mode collapse.
These shapes also exhibit poor compliance. \citet{sanu2024neural_good_bad_ugly} demonstrated that neural networks can more effectively explore the solution space of topology optimization (TO) problems. We therefore hypothesize that inducing higher diversity, either at initialization or as a constraint, may promote further exploration.

\section{Geometric constraints}
\label{app:constraints}
We formulate geometric constraints analogously to GINNs in Table \ref{tab:constraints} for density representations.
There are two important differences when changing the shape representation from a signed distance function (SDF) to a density function.\newline
First, the level set $\partial \Omega_{\tau}$ changes from 
$ \partial \Omega_{\tau} = \left\{x \in \mathbb{R}^{d_x} | f_{\theta} = 0 \right\}$ to $ \partial \Omega_{\tau} = \left\{x \in \mathbb{R}^{d_x} | f_{\theta} = 0.5 \right\}$.
Second, for a SDF the shape $\Omega_{tau}$ is defined as the sub-level set $\Omega_{\tau} = \left\{x \in \mathbb{R}^{d_x} | f_{\theta} \leq 0 \right\}$ , whereas for a density it is the super-level set $\Omega_{\tau} = \left\{x \in \mathbb{R}^{d_x} | f_{\theta} \geq 0 \right\}$.

\def\normal{ \frac{\nabla f(x)}{||\nabla f(x)||}}
\renewcommand{\arraystretch}{1.5} %
\begin{table*}
    \small
    \caption{
        Geometric constraints are derived from GINNs.
        The shape $\Omega$ and its boundary $\partial\Omega$ are implicitly defined by the level set $\tau$ of the function $f$.
        The shape must reside within the \emph{design region} $\mathcal{E}\subseteq\mathcal{X}$ and adhere to the \emph{interface} $\mathcal{I}\subset\mathcal{E}$ with a specified \emph{normal} $\bar{n}(x)$.
        }
    \centering
    \begin{tabular}{l|l|l|l}
        & Set constraint $c_i(\Omega)$ & Function constraint & Constraint violation $c_i(f)$ \\
        \hline
        Design region 
            & $\Omega \subset \mathcal{E}$ 
            & $f(x) < \tau \ \forall x \notin \mathcal{E} $
            & $\int_{\mathcal{X}\setminus\mathcal{E}} \left[ \operatorname{max}(0, f(x) - \tau) \right]^2 dx$ \\
        Interface 
            & $\partial\Omega \supset \mathcal{I}$
            & $f(x) = \tau \ \forall x \in \mathcal{I}$
            & $\int_\mathcal{I} \left[ f(x) - \tau \right]^2 dx$ \\
        Prescribed normal
            & $n(x)=\bar{n}(x) \ \forall x \in \mathcal{I}$ 
            & $\normal = \bar{n}(x) \ \forall x \in \mathcal{I}$
            & $\int_\mathcal{I} \left[ \normal - \bar{n}(x)\right]^2 dx$ \\
    \end{tabular}
    
    \label{tab:constraints}
\end{table*}

\section{Diversity}

\subsection{Diversity on the volume}
\label{subsec:volume_dissimilarity}

As noted by \citet{GINNs}, one can define a dissimilarity loss as the $L^p$ function distance
\begin{equation}
d(\Omega_i, \Omega_j) = \sqrt[p]{\int_\mathcal{X} (f_i(x) - f_j(x))^p dx}
\end{equation}
We choose $L^1$, to not overemphasize large differences in function values.
Additionally, we show in the next paragraph that for $L^1$ and for the extreme case where $f(x) \in \{0, 1\}$ this is equal to the Union minus Intersection of the shapes: 

\begin{equation}
    \int_\mathcal{X} |f_i(x) - f_j(x)| dx = \text{Vol}(\Omega_i \cup \Omega_j) - \text{Vol}(\Omega_i \cap \Omega_j)
\end{equation}

\subsection{\texorpdfstring{$L^1$ distance on neural fields resembles Union minus Intersection}{L1 distance on neural fields resembles Union minus Intersection}}
To derive that the distance metric 

\begin{equation}
     d(\Omega_i, \Omega_j) = \sqrt[p]{\int_\mathcal{X} (f_i(x) - f_j(x))^p dx}
\end{equation}

for \( p=1 \) and \( f_i, f_j \in \{0,1\} \) corresponds to the union minus the intersection of the shapes, we consider the following cases:

\begin{table}[h!]
    \centering
    \begin{tabular}{c|c|c}
    \( f_i(x) \) & \( f_j(x) \) & \( |f_i(x) - f_j(x)| \) \\
    \hline
    1 & 1 & 0 \\
    1 & 0 & 1 \\
    0 & 1 & 1 \\
    0 & 0 & 0 \\
    \end{tabular}
    \caption{Function distance if the function only has binary values.}
    \label{tab:binary}
\end{table}

From the table, we observe that the integrand \( |f_i(x) - f_j(x)| \) is 1 when \( x \) belongs to one shape but not the other, and 0 when \( x \) belongs to both or neither.
Thus, the integral \( \int_\mathcal{X} |f_i(x) - f_j(x)| dx \) sums the volumes where \( x \) is in one shape but not the other, which is precisely the volume of the union of \( \Omega_i \) and \( \Omega_j \) minus the volume of their intersection.
It follows that 

\begin{equation}
    d(\Omega_i, \Omega_j) = \int_\mathcal{X} |f_i(x) - f_j(x)| dx = \text{Vol}(\Omega_i \cup \Omega_j) - \text{Vol}(\Omega_i \cap \Omega_j) 
\end{equation}.

\subsection{Diversity on the boundary via differentiable chamfer discrepancy}
\label{subsec:chamfer_diversity}

We continue from Equation \ref{eq:level_set}:
\begin{align}
    \frac{\partial L}{\partial \theta} &= \frac{\partial L}{\partial x} \frac{\partial x}{\partial y} \frac{\partial y}{\partial \theta} \nonumber \\ 
    &= \frac{\partial L}{\partial x} \frac{\nabla_x f_\theta}{|\nabla_x f_\theta|^2} \frac{\partial y}{\partial \theta}
\end{align}
The center term $\frac{\nabla_x f_\theta}{|\nabla_x f_\theta|^2}$ and the last term  $\frac{\partial y}{\partial \theta}$ can be obtained via automatic differentiation.
For the first term, we derive $\frac{\partial L}{\partial x}$, where $L$ is the one-sided chamfer discrepancy $\text{CD}(\partial \Omega_1, \partial \Omega_2)$.
\begin{align}
\frac{\partial}{\partial x}  \text{CD}(\partial \Omega_1, \partial \Omega_2)
&= \frac{\partial }{\partial x} \frac{1}{\left| \partial \Omega_1 \right|} \sum_{x \in \partial \Omega_1} \min_{y \in \partial \Omega_2} ||x-y||_2
\\
&= \frac{1}{\left| \partial \Omega_1 \right|} \min_{y \in \partial \Omega_2} \frac{x - y}{\|x - y\|}
\end{align}
This completes the terms in the chain rule.

\subsection{Finding surface points}
We detail the algorithm to locate boundary points of implicit shapes defined by a neural density field, Algorithm \ref{alg:boundary_points}. \\
On a high level, the algorithm first identifies points inside the boundary where neighboring points lie on opposite sides of the level set.
Subsequently, it employs binary search to refine these boundary points.
The process involves evaluating the neural network to determine the signed distance or density values, which are then used to iteratively narrow down the boundary points.
We find that 10 binary steps suffice to reach the boundary sufficiently close.

For the 3D jet engine bracket, we additionally follow \citet{GINNs} and exclude surface points which are within some $\epsilon$ region of the interface.

\begin{algorithm}[ht]
\caption{Find Boundary Points with Binary Search}
\label{alg:boundary_points}
\begin{algorithmic}
\STATE {\bfseries Input:}
regular point grid $x_i$, level set $l$, neural density field $f_\theta$, binary search steps $T$
\STATE Find points $x_i^\text{in} \in x_i$ for which $f(x_i^\text{in}) > l$
\STATE Find neighbor points $(x_i^\text{out})_i \in x_i$ to $x_i^\text{in}$ $f(x_\text{out}) < l$ \COMMENT{ for 2D/3D we check 4/6 neighbors respectively}
\STATE Refine pairs $(x_\text{in}, x_\text{out})_i$ via $T$ binary search steps
\STATE Return $((x_\text{in} + x_\text{out})/2)_i$
\end{algorithmic}
\end{algorithm}

\section{Topology optimization}
\label{app:method_details}

\subsection{Problem definitions}
\label{app:problem_defs}

\paragraph{Messerschmitt-Bölkow-Blohm beam (2D).}
The MBB beam is a common benchmark problem in TO and is depicted in Figure \ref{fig:beam_def}. The problem describes a beam fixed on the lower right and left edges with a vertical force $F$ applied at the center. As the problem is symmetric around $x=L/2$, we follow \citet{multi_TO_Papadopoulos_2021} and only optimize the right half. 
\newline
The concrete dimensions of the beam are $H=1, L=6$, and the force points downwards with $F=1$. 

\paragraph{Cantilever beam (2D).}
The cantilever beam is illustrated in Figure \ref{fig:cantilever_def}. The problem describes a beam fixed on the left-hand side, and two forces $F_1$, $F_2$ are applied on the right-hand side. \newline
The dimensions are $H=1, L=1.5, h=0.1$ and the forces $ F_1 = F_2 = 0.5$.

\paragraph{Jet engine bracket (3D).}
We apply TOM to a challenging 3D task, namely the optimization of a jet engine bracket as defined by \citet{jeb}.
The design region for this problem is enclosed by a freeform surface mesh (see Figure \ref{fig:simjeb_combined}).
The load case is depicted in Figure \ref{fig:simjeb_combined} in which a diagonal force pulls in positive x and positive z direction.

\subsection{Smoothing and contrast filtering}
\label{subsec:density_filters}
Filtering is a general concept in topology optimization that aims to reduce artifacts and improve convergence.
\emph{Helmholtz PDE filtering} is a smoothing filter similar to Gaussian blurring, but easier to integrate with existing finite element solvers. %
By solving a Helmholtz PDE, the material density $\rho$ is smoothened to prevent checkerboard patterns, which is typical for TO \citep{lazarov2011filters}.
\emph{Heaviside filtering} is a type of contrast filter, which enhances the distinction between solid and void regions.
The Heaviside filter function, as defined in Equation~\ref{eq:heaviside}, equals the sigmoid function up to scaling of the input.
$\beta$ is a parameter controlling the sharpness of the transform, similar to the inverse temperature of a classical softmax (higher beta means a closer approximation to a true Heaviside step function).
Note that in contrast to the Helmholtz PDE filter, the Heaviside filter is not volume preserving. Therefore, the volume constraint has to be applied to the modified output.

\begin{equation}
\label{eq:heaviside}
H(x, \beta) = 0.5 + \frac{\tanh(\beta(x - 0.5))}{2 \tanh(0.5 \beta)} \
\end{equation}

\subsection{Post-processing}
\label{app:postprocess}

In general, the post-processing of TO solutions heavily depends on the downstream task and the specific manufacturing requirements, e.g., additive vs subtractive manufacturing. Here, we demonstrate two simple methods to post-process TOM solutions:

\paragraph{Method \textbf{A}} is a simple yet standard image-based approach that removes disconnected components (floaters) and performs morphological closing to remove small artifacts (holes).

\paragraph{Method \textbf{B}} is a fine-tuning approach that performs a few iterations of classical TO updates (5\% of a full run, learning rate reduced by $\times 10$ compared to the standard FeniTop settings) on the TOM output. This approach also removes the artifacts. This illustrates how TOM extends, not competes, with classical TO.

\begin{figure}[hb!]
    \centering
    \begin{subfigure}{0.8\linewidth}
        \centering
            \includegraphics[width=\linewidth]{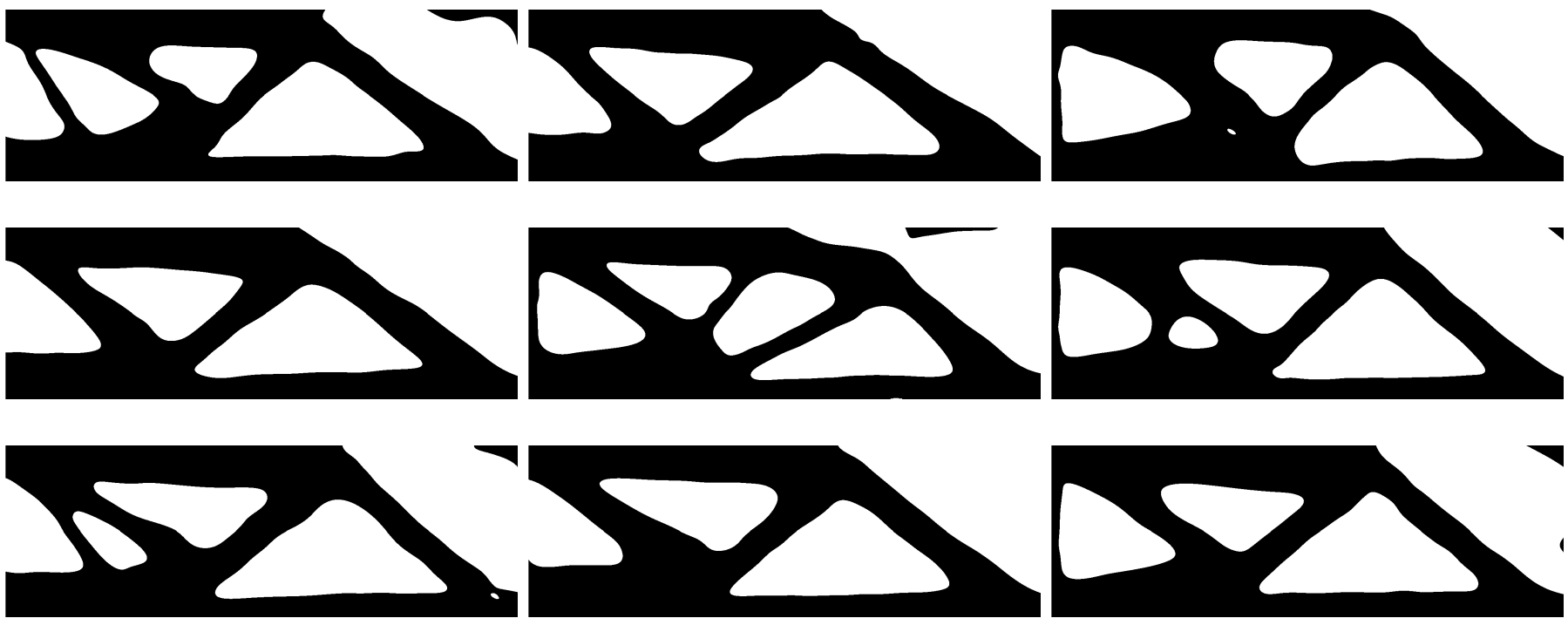}
        \caption{Output of TOM model.}
        \label{fig:TOM_output_initial}
    \end{subfigure}
    
    \begin{subfigure}{0.8\linewidth}
        \centering
        \includegraphics[width=\linewidth]{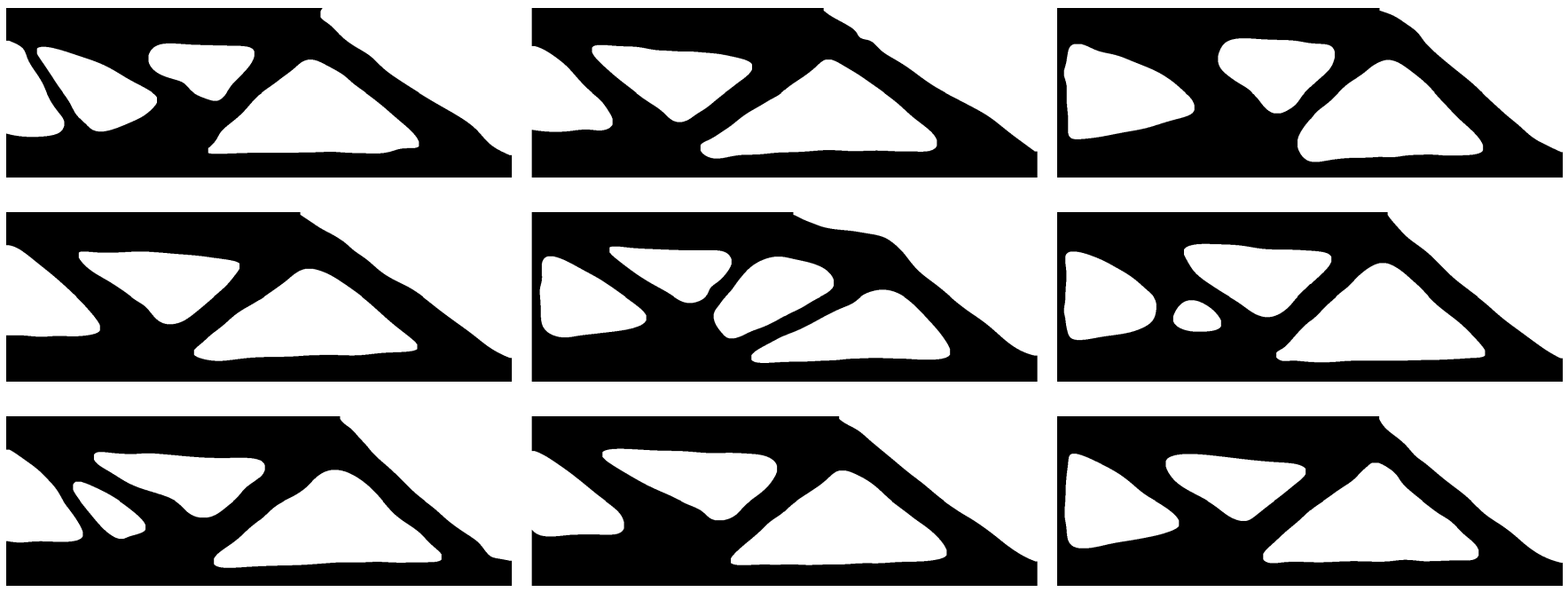}
        \caption{Method \textbf{A}: Simple post-processing using OpenCV to remove artifacts.}
        \label{fig:opencv_pp}
    \end{subfigure}

    \begin{subfigure}{0.8\linewidth}
        \centering
        \includegraphics[width=\linewidth]{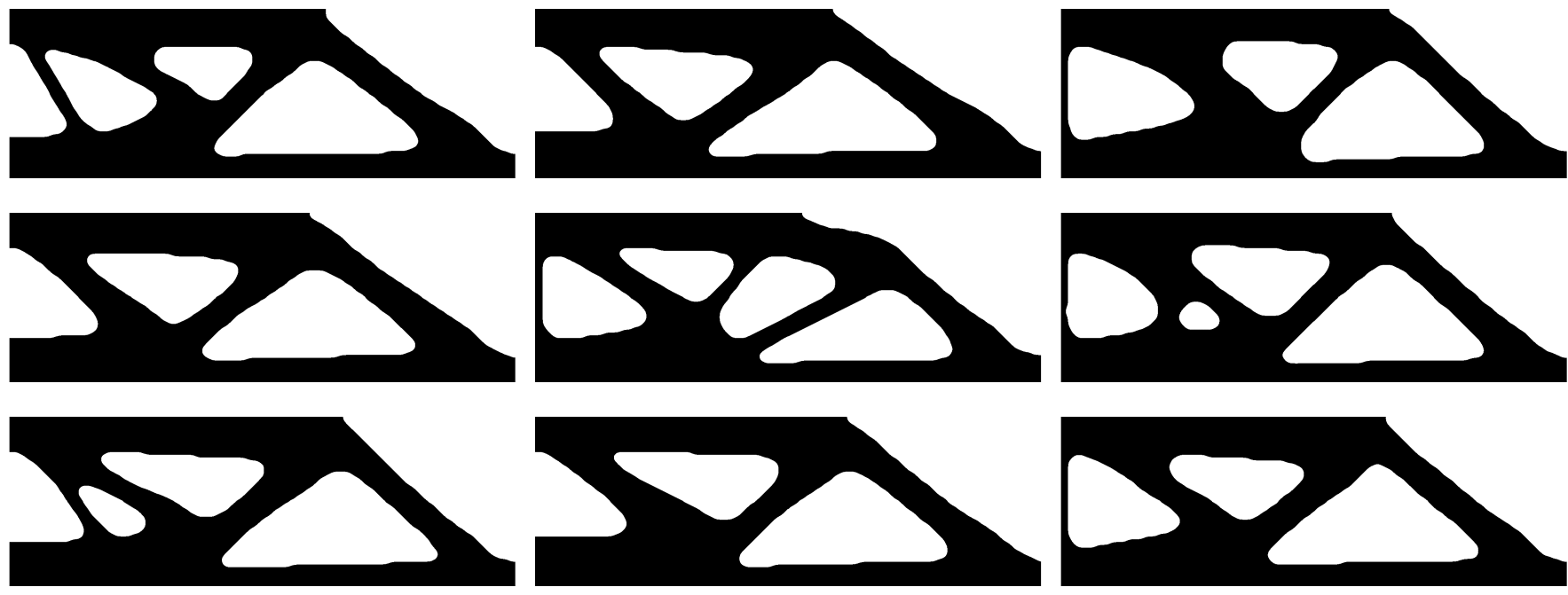}
        \caption{Method \textbf{B}: Finetune by running a few FeniTop iterations to convergence (5\% of total iterations).}
        \label{fig:topopt_pp}
    \end{subfigure}
    
    \caption{\textbf{Post-processing: } Showcasing 2 possible post-processing steps to remove artifacts from the TOM solutions.}
    \label{fig:pp_mbb_beam}
\end{figure}

We depict a version of an MBB beam with many floaters and its post-processed version in Figure \ref{fig:pp_mbb_beam}.

\end{document}